\documentclass[lettersize,journal]{IEEEtran}
\usepackage{amsmath}
\usepackage{amsfonts}
\usepackage{algorithm}
\usepackage{array}
\usepackage[caption=false,font=footnotesize]{subfig}
\usepackage{textcomp}
\usepackage{stfloats}
\usepackage{url}
\usepackage{verbatim}
\usepackage{graphicx}
\usepackage[citecolor=blue, colorlinks]{hyperref}
\usepackage{cite}
\usepackage{booktabs} 
\usepackage{tikz}
\usepackage{bm}
\usepackage{algpseudocode}
\usepackage{amssymb}
\def\BibTeX{{\rm B\kern-.05em{\sc i\kern-.025em b}\kern-.08em
    T\kern-.1667em\lower.7ex\hbox{E}\kern-.125emX}}
\usepackage{balance}
\usepackage{xcolor}
\usepackage{soul}

\begin{document}

\title{Enhancing Underwater Light Field Images via Global Geometry-aware Diffusion Process}

\author{Yuji Lin, Qian Zhao, \textit{Member}, \textit{IEEE}, Zongsheng Yue, \textit{Member}, \textit{IEEE}, Junhui Hou, \textit{Senior Member}, \textit{IEEE}, and Deyu Meng, \textit{Member}, \textit{IEEE}
        % <-this % stops a space
\thanks{This work was supported in part by the National Key R\&D Program
of China under Grant 2024YFA1012201, in part by the National NSFC projects under Grants 62422118, 12471485, and 62331028, and in part by the Hong Kong Research Grants Council under Grant 11218121. (\textit{Corresponding author: Junhui Hou and Qian Zhao}.)}
\thanks{Yuji Lin and Zongsheng Yue are with the School of Mathematics and Statistics, Xi’an Jiaotong University, Xi’an, Shaanxi 710049, China (email: linlos1234@gmail.com, zsyue@xjtu.edu.cn).}
\thanks{Qian Zhao is with the School of Mathematics and Statistics and the Ministry of Education Key Lab of Intelligent Networks and Network Security, Xi’an Jiaotong University, Shaanxi 710049, China (email: timmy.zhaoqian@mail.xjtu.edu.cn).

Junhui Hou is with the Department of Computer Science, City University of Hong Kong, Hong Kong SAR (email: jh.hou@cityu.edu.hk).  

Deyu Meng is with the School of Mathematics and Statistics and the Ministry of Education Key Laboratory of Intelligent Networks and Network Security, Xi’an Jiaotong University, Xi’an, Shaanxi 710049, China, and also with the Macao Institute of Systems Engineering, Macau University of Science and Technology, Taipa, Macao (email: dymeng@mail.xjtu.edu.cn).}% <-this % stops a space
}

% The paper headers
\markboth{IEEE Transactions on Image Processing}%
{Shell \MakeLowercase{\textit{et al.}}: A Sample Article Using IEEEtran.cls for IEEE Journals}

% \IEEEpubid{0000--0000/00\$00.00~\copyright~2021 IEEE}

\maketitle

\begin{abstract}
This work studies the challenging problem of acquiring high-quality underwater images via 4-D light field (LF) imaging. 
To this end, we propose GeoDiff-LF, a novel diffusion-based framework built upon SD-Turbo to enhance underwater 4-D LF imaging by leveraging its spatial-angular structure.
GeoDiff-LF consists of three key adaptations: (1) a modified U-Net architecture with convolutional and attention adapters to model geometric cues, (2) a geometry-guided loss function using tensor decomposition and progressive weighting to regularize global structure, and (3) an optimized sampling strategy with noise prediction to improve efficiency.
By integrating diffusion priors and LF geometry, GeoDiff-LF effectively mitigates color distortion in underwater scenes. Extensive experiments demonstrate that our framework outperforms existing methods across both visual fidelity and quantitative performance, advancing the state-of-the-art in enhancing underwater imaging. The code will be publicly available at \url{https://github.com/linlos1234/GeoDiff-LF}.
\end{abstract}


\section{Introduction}
\label{sec-intro}

\IEEEPARstart{U}{nderwater} imaging plays a pivotal role in the burgeoning marine research and ocean economy~\cite{gu2023underwater}. However, the complex underwater environment, characterized by diverse water compositions and particulate matter, introduces significant challenges to image quality. The absorption, refraction, and scattering of light in water lead to severe degradation of underwater images, manifesting as color distortion, reduced brightness, low contrast, and foggy blurring~\cite{akkaynak2017space}. Moreover, the degree of degradation is exacerbated by increasing water depth. Consequently, mitigating these degradations is critical to improving the quality of underwater imaging and enhancing the performance of downstream applications.

A wide range of methods~\cite{chiang2011underwater,wang2017single,li2019underwater, liu2022adaptive, akkaynak2018revised, li2021underwater} have been developed for underwater image enhancement, including both traditional algorithms and deep learning approaches. 
Among these, degradation-aware methods~\cite{wang2017single,li2019underwater, akkaynak2018revised, li2021underwater} explicitly model the underlying physical processes of underwater imaging, consistently improving performance by incorporating geometric information, such as scene depth. Recently, diffusion model~\cite{ho2020denoising, Rombach_2022_CVPR} have emerged as powerful generative prior for image restoration~\cite{ren2023multiscale, chen2023hierarchical, jiang2023low, qing2024diffuie, du2025uiedp, yue2023resshift, hou2023global}. By integrating task-specific guidance into the intricate process, these approaches~\cite{yue2023resshift, hou2023global, yue2024difface} further improve denoising accuracy and restoration fidelity, offering a promising avenue for underwater image enhancement.

4-D light field (LF) imaging~\cite{ng2006digital}, leveraging its unique capture mechanism, simultaneously records the reflected light intensity and geometric information from multiple viewpoints. This is typically achieved by a multi-camera configuration, which allows the acquisition of dense angular information. Compared to conventional 2-D RGB imaging, the geometric structure captured in 4-D LF imaging significantly enhances underwater image restoration~\cite{lin2025enhancing}. Furthermore, the fusion of multi-view light rays effectively mitigates the low-light condition and suppresses noise. This geometric structure provides a robust foundation for addressing the complex degradations inherent in underwater environments, making 4-D LF imaging a powerful tool for advancing restoration techniques.

Although the rich information embedded in 4-D LF offers substantial potential for underwater imaging, its complex image structure also poses significant challenges for effective processing. To address this, we seek to extend the powerful image priors in diffusion models~\cite{ho2020denoising, Rombach_2022_CVPR} to 4-D LF-based underwater imaging. Nevertheless, directly applying diffusion models to 4-D LF underwater images faces three critical obstacles: (1) existing diffusion models are predominantly trained on 2-D RGB images, leading to an inherent dimensionality mismatch when handling 4-D LFs; (2) the geometric structure of 4-D LFs may be disrupted during both the forward and reverse diffusion processes; and (3) the computational cost of deploying complex diffusion models on high-dimensional LF data could be extremely high, which hinders efficient processing.

In this work, we propose GeoDiff-LF, a global geometry-aware diffusion framework for underwater LF image enhancement that effectively addresses the aforementioned challenges. Our method adaptively tailors the diffusion backbone to model spatial-angular representations and introduces global geometry regularization into the denoising process to steer the diffusion model toward intrinsic LF structures. An efficient noise prediction strategy is further designed to achieve restoration with fewer steps. By tightly incorporating the diffusion prior with the unique geometric properties of 4-D LFs, GeoDiff-LF achieves superior performance in enhancing underwater imaging. The key contributions of this paper are summarized as follows:
\begin{itemize}
    \item We adapt the diffusion architecture to efficiently handle 4-D LFs, effectively exploiting their inherent geometric and multi-view information to acquire high-quality underwater images. 
    \item We design a global geometry regularization that aligns the tensor-level intrinsic features across the spatial and angular dimensions of 4-D LFs, thereby enhancing geometric consistency and content coherence throughout the denoising process. 
    \item An efficient noise prediction mechanism is proposed by introducing an auxiliary noise map predictor, allowing the model to bypass early reconstruction stages and achieve high-quality restoration with significantly fewer reversion steps.
\end{itemize}

The remainder of this paper is organized as follows. In Section~\ref{sec:rw}, we briefly review the related work of image restoration methods, including LF-based and diffusion-based methods. In Section~\ref{sec:pre}, we formulate the forward and backward processes of diffusion models. In Section~\ref{sec:method}, we propose the framework of our GeoDiff-LF in detail. Experimental results are presented in Section~\ref{sec:exp}.
Finally, we conclude this paper and discuss the future work in Section~\ref{sec:c}.

\section{Related Work}
\label{sec:rw}
\subsection{Light Field Image Restoration}
Existing research on LF image restoration encompasses a wide range of low-level tasks, including super-resolution~\cite{yeung2018light, wang2022disentangling, cong2023exploiting, liang2023learning, mao2025learning, mao2025deep, liu2025learning, liu2024light}, denoising~\cite{guo2021deep, wang2023multi, lyu2024probabilistic}, low-light enhancement~\cite{lamba2020harnessing, zhang2023lrt, lyu2024enhancing}, and deraining~\cite{ding2021rain, yan2023rain, lyu2024rainyscape}. 

Researchers prioritize the spatial-angular information inherent in 4-D LFs. They have employed various approaches to leverage this information for restoring low-quality images. One important class of methods involves using convolutional networks to extract the required multidimensional features. 
Yeung~\emph{et~al.}~\cite{yeung2018light} proposed to use spatial-angular separable convolutions as an approximation of 4-D convolution for more computationally and memory-efficient extraction of spatial-angular joint features.
Lyu~\emph{et~al.}~\cite{lyu2024enhancing} designed a pseudo-explicit feature interaction module that performs convolution on all domains of four LF dimensions to exploit redundant information in LF images.
Wang~\emph{et~al.}~\cite{wang2023multi} introduced a multi-stream network to progressively extract features from different views of LF images.
Mao~\emph{et~al.}~\cite{mao2025deep} proposed a baseline that synthesizes dense novel views from sparse multi-view inputs using an adaptive alignment module and a multi-level feature decoupling module to exploit angular consistency and hierarchical cues.
Liu~\emph{et~al.}~\cite{liu2025learning} proposed a spatial-angular super-resolution framework that combines implicit neural representation learning with explicit detail-enhanced modules to jointly reconstruct high-resolution spatial and angular LF images.
Ding~\emph{et~al.}~\cite{ding2021rain} proposed the insight that depth information can help distinguish rain streaks in epipolar plane images of LF images and leverage this property to design a GAN-based framework for rain streak removal.
The long-range dependency of Transformers facilitates the modeling of global spatial-angular correlations. Developers explore self-attention mechanisms across different dimensions~\cite{cong2023exploiting, liang2023learning}, which establish dependencies by leveraging representational differences across all dimensions of 4-D LFs. 
Zhang~\emph{et~al.}~\cite{zhang2023lrt} designed
a transformer-based model tailored for learning different components of low-light LF images.
Chao~\emph{et~al.}~\cite{chao2024bigepit} scaled EPIT~\cite{liang2023learning} called BigEPIT with more depth information and model width to improve model capability.

However, few studies~\cite{lin2025enhancing, zhou2025lfuid} have attempted underwater LF image enhancement, leaving this vital topic yet underexplored. 
Lin~\emph{et~al.}~\cite{lin2025enhancing} first introduced LF imaging for enhancing underwater imaging. They conducted a vivid pairwise LF image dataset comprising 75 underwater scenes and designed a progressive, mutually reinforcing framework for underwater image enhancement and depth estimation.
Zhou~\emph{et~al.}~\cite{zhou2025lfuid} captured a real LF image dataset with large-scale data and proposed a model-based approach via angular correspondence in LF images to remove underwater scattering effects and enhance underwater images.

\begin{algorithm}[t]
\caption{Training process of DDPMs}
\label{alg:forward}
\begin{algorithmic}[1]
\State \textbf{Repeat}
\State $\mathbf{X}_0\sim q(\mathbf{X}_0)$
\State $t\sim \text{Uniform}\left(\{1,2,\dots,T\}\right)$
\State $\epsilon\sim \mathcal{N}(0,\bm{I})$
\State Perform a gradient descent step on 
\State ~~~~$\nabla_{\theta} \left\| \epsilon-\epsilon_{\theta}(\mathbf{X}_t,\mathbf{Y}_0,t) \right\|_2^2$ 
\State \textbf{Until} converged
\end{algorithmic}
\end{algorithm}
\begin{algorithm}[t]
\caption{Inference process of DDPMs}
\label{alg:reverse}
\begin{algorithmic}[1]
\State $\mathbf{X}_T\sim \mathcal{N}(0,\bm{I})$
\State \textbf{For} $t=T,\dots,1$
\State ~~~~$\mathbf{Z}_{t-1}\sim \mathcal{N}(0,\bm{I})$ if $t>1$, else $\mathbf{Z}_{t-1}=0$
\State ~~~~$\mathbf{X}_{t-1}=g_{\theta}(\mathbf{X}_{t},\mathbf{Y}_0,t) + \sigma_{t}\mathbf{Z}_{t-1}$, where $g_{\theta}$ is defined in Eq.~\eqref{eq:g-theta}
\State \textbf{End for}
\State \textbf{Return} $\mathbf{X}_0$
\end{algorithmic}
\end{algorithm}

\subsection{Diffusion Models for Image Restoration}
In recent years, diffusion models have significantly progressed in handling low-level vision tasks of 2-D RGB images, such as super-resolution~\cite{yue2023resshift, saharia2022image, yue2025arbitrary}, deblurring~\cite{ren2023multiscale, chen2023hierarchical}, low-light enhancement~\cite{jiang2023low, hou2023global}, and underwater enhancement~\cite{qing2024diffuie, du2025uiedp}. 
Saharia~\emph{et~al.}~\cite{saharia2022image} pioneered the application of diffusion models to image super-resolution, incorporating low-resolution images as conditional information into the denoising process. 
Yue~\emph{et~al.} proposed ResShift~\cite{yue2023resshift}, utilizing a Markov chain with residual shifting between high- and low-resolution images to enhance transition efficiency, achieving impressive performance with only 15 sampling steps. 
Hou~\emph{et~al.}~\cite{hou2023global} developed a global structure-aware diffusion model to enhance low-light imaging, integrating structural priors into the denoising process to effectively preserve global image coherence and improve illumination quality.
Chen~\emph{et~al.}~\cite{chen2023hierarchical} proposed a hierarchical integration module that fuses prior knowledge into a regression-based model at multiple scales, enabling better generalization in complex and blurry scenarios.
Qing~\emph{et~al.}~\cite{qing2024diffuie} proposed DiffUIE, a diffusion model that leverages latent global priors, employing feature-level guidance to mitigate noise and color distortion for underwater images.
Du~\emph{et~al.}~\cite{du2025uiedp} suggested to boost underwater image enhancement with a pre-trained diffusion prior.
Zhu~\emph{et~al.}~\cite{zhu2025learning} proposed a novel differential equation-based diffusion framework to restore low-quality images, via a reinforcement learning-based ODE trajectory augmentation algorithm.
Although these methods achieved promising results on 2-D RGB images, they cannot be directly applied to 4-D LFs due to an inherent dimensionality mismatch.

Diffusion models have also demonstrated considerable capabilities in addressing other multidimensional data~\cite{rui2024unsupervised, xing2024simda, huang2024epidiff}.
Rui~\emph{et al.}~\cite{rui2024unsupervised} developed a low-rank diffusion model for hyperspectral pan sharpening, effectively combining the strengths of pre-trained diffusion models with the enhanced generalization capabilities of Bayesian approaches.
Xing~\emph{et al.}~\cite{xing2024simda} inserted lightweight spatial and temporal adapters into a strong T2I model, which preserved the expressive power of the large-scale model and effectively handled temporal-spatial correlations.
Huang~\emph{et al.}~\cite{huang2024epidiff} integrated a geometric epipolar attention block into the frozen diffusion model to enable cross-view interaction among feature maps of neighboring views for multi-view reconstruction.

Although a few existing studies~\cite{zhang2024lfir,gao2025diff,chao2025lfsr} have attempted to adapt 4-D LF-based diffusion models by modifying the original network architecture and training from scratch, they exhibit limited performance and face significant computational efficiency challenges.
Zhang~\emph{et~al.}~\cite{zhang2024lfir} added view-dependent noise to LF images and converted them into diffusion models for denoising.
Gao~\emph{et~al.}~\cite{gao2025diff} designed a position-aware warping condition scheme and replaced the original 2-D convolutions with a disentangling mechanism~\cite{wang2022disentangling}.
We draw inspiration from diffusion models developed for other image restoration applications, as aforementioned, while specifically designing the network modules and training strategies to accommodate the unique characteristics of 4-D LFs.

\section{Preliminaries}
\label{sec:pre}
To elucidate the theoretical foundation of our approach, we adopt the Denoising Diffusion Probabilistic Model (DDPM) framework~\cite{ho2020denoising} as the basis for GeoDiff-LF. The DDPM indeed operates as a Markov chain over $T$ steps, with the forward process defined by a Gaussian transition kernel:
\begin{equation}
    q(\mathbf{X}_t|\mathbf{X}_{t-1})=\mathcal{N}\left(\mathbf{X}_t; \sqrt{1-\beta_t}\mathbf{X}_{t-1}, ~\beta_t I\right),
\end{equation}
where $\beta_t$ is a pre-defined hyper-parameter controlling the variance schedule. This form enables a closed-form marginal distribution $q(\mathbf{X}_t|\mathbf{X}_0)$ at an arbitrary timestep $t$, namely
\begin{equation}
    q(\mathbf{X}_t|\mathbf{X}_0)=\mathcal{N}\left(\mathbf{X}_t; \sqrt{\bar{\alpha}_t}\mathbf{X}_{t-1}, ~(1-\bar{\alpha}_t) I\right),
\end{equation}
where $\bar{\alpha}_t=\prod_{s=1}^t \alpha_s$, $\alpha_s=1-\beta_s$. The reverse process aims to reconstruct a high-quality image from an initial random noise map $\mathbf{X}_T \sim \mathcal{N}(0,\bm{I})$, formulated as:
\begin{equation}
    \mathbf{X}_{t-1} = g_{\theta}(\mathbf{X}_t, t)+\sigma_t \mathbf{Z}_{t-1}, ~t=T,\cdots,1,
\end{equation}
where
\begin{equation}
    g_\theta(\mathbf{X}_t, t)=\frac{1}{\sqrt{\alpha_t}}\left(\mathbf{X}_t-\frac{1-\alpha_t}{\sqrt{1-\bar{\alpha}_t}} \epsilon_\theta (\mathbf{X}_t, t)\right),
\end{equation}
and $\epsilon_\theta (\mathbf{X}_t,~t)$ is a denoising network parameterized by $\theta$. The noise term $\mathbf{Z}_t$ satisfies $\mathbf{Z}_0=0$ and $\mathbf{Z}_t \sim \mathcal{N}(0,\bm{I})$ for $t=1,\cdots,T-1$. For image restoration, the denoiser is generally conditioned with the low-quality image $\mathbf{Y}_0$:
\begin{equation}
    \mathbf{X}_{t-1} = g_{\theta}(\mathbf{X}_t,\mathbf{Y}_0,t)+\sigma_t \mathbf{Z}_{t-1}, ~t=T,\cdots,1,
    \label{eq:inter-recon}
\end{equation}
where
\begin{equation}
    g_\theta(\mathbf{X}_t,\mathbf{Y}_0,t)=\frac{1}{\sqrt{\alpha_t}}\left(\mathbf{X}_t-\frac{1-\alpha_t}{\sqrt{1-\bar{\alpha}_t}} \epsilon_\theta (\mathbf{X}_t,~\mathbf{Y}_0,~t)\right).
    \label{eq:g-theta}
\end{equation}

\begin{figure*}[t]
\centering
\includegraphics[width=1\linewidth]{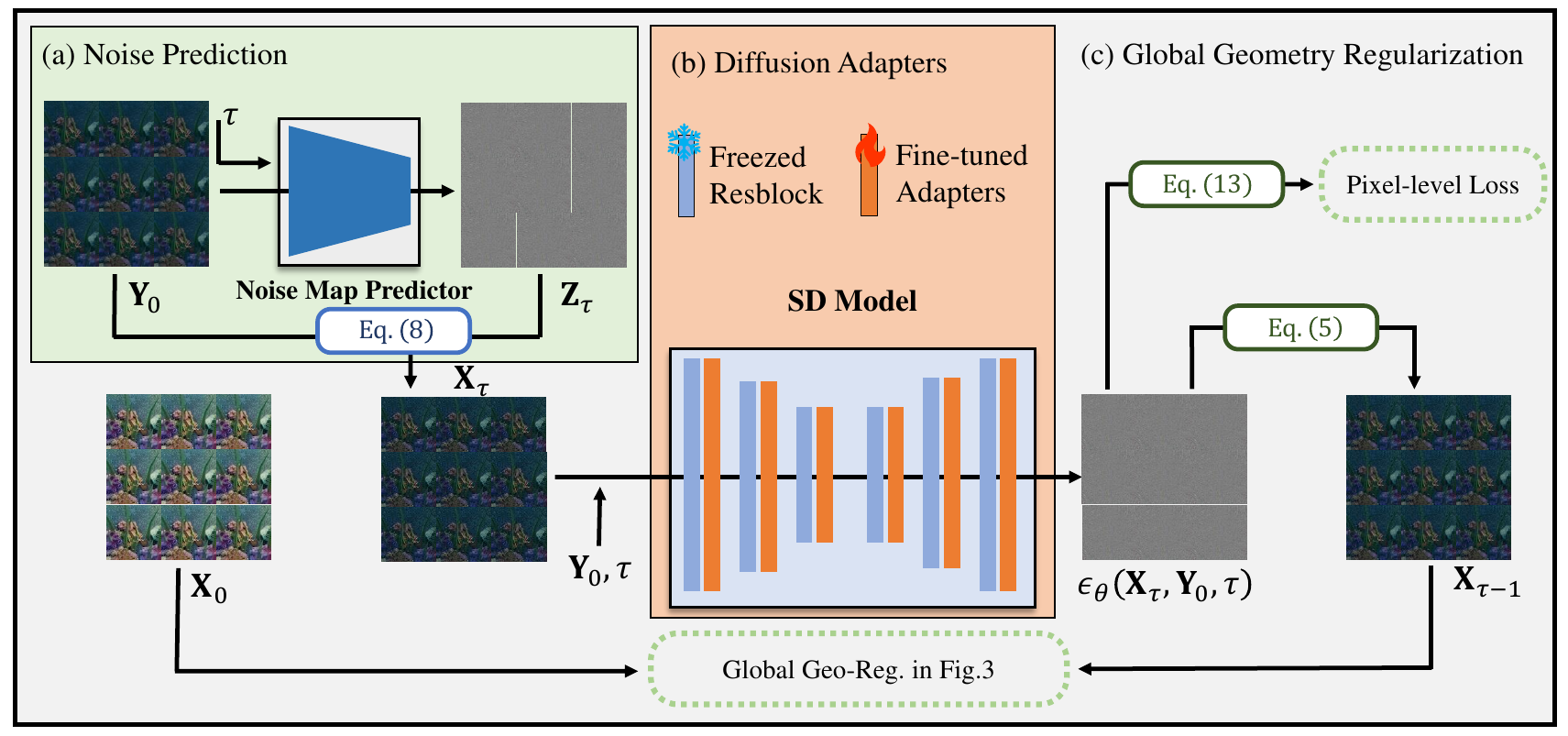}
\caption{The whole pipeline of the proposed GeoDiff-LF. For the training phase, we first employ a noise map predictor to generate a noisy sample $\mathbf{X}_{\tau}$ from the underwater LF image $\mathbf{Y}_0$ and timestep $\tau$, where $\tau(\tau<T)$ is the start timestep. We then feed this noisy sample, together with the corresponding underwater image $\mathbf{Y}_0$ and noise level $\bar{\alpha}_{\tau}$, into our adapter-augmented U-Net $\epsilon(\cdot)$ to predict the added noise $\epsilon$. To thoroughly capture global geometric structure, we construct an intermediate reconstruction $\mathbf{X}_{\tau-1}$ from the predicted noise $\epsilon_\theta (\mathbf{X}_{\tau},\mathbf{Y}_0,\tau)$ and undergo additional regularization throughout the restoration process. While testing LF underwater images, we start at timestep $\tau$ to generate the noisy sample $\mathbf{X}_{\tau}$, then predict the corresponding noise to obtain the enhanced result.}
\label{fig-pipeline}
\end{figure*}

\begin{figure*}[t]
    \centering
    \begin{tikzpicture}
        \node[anchor=south west,inner sep=0] (image) at (0,0) {\includegraphics[width=0.8\textwidth]{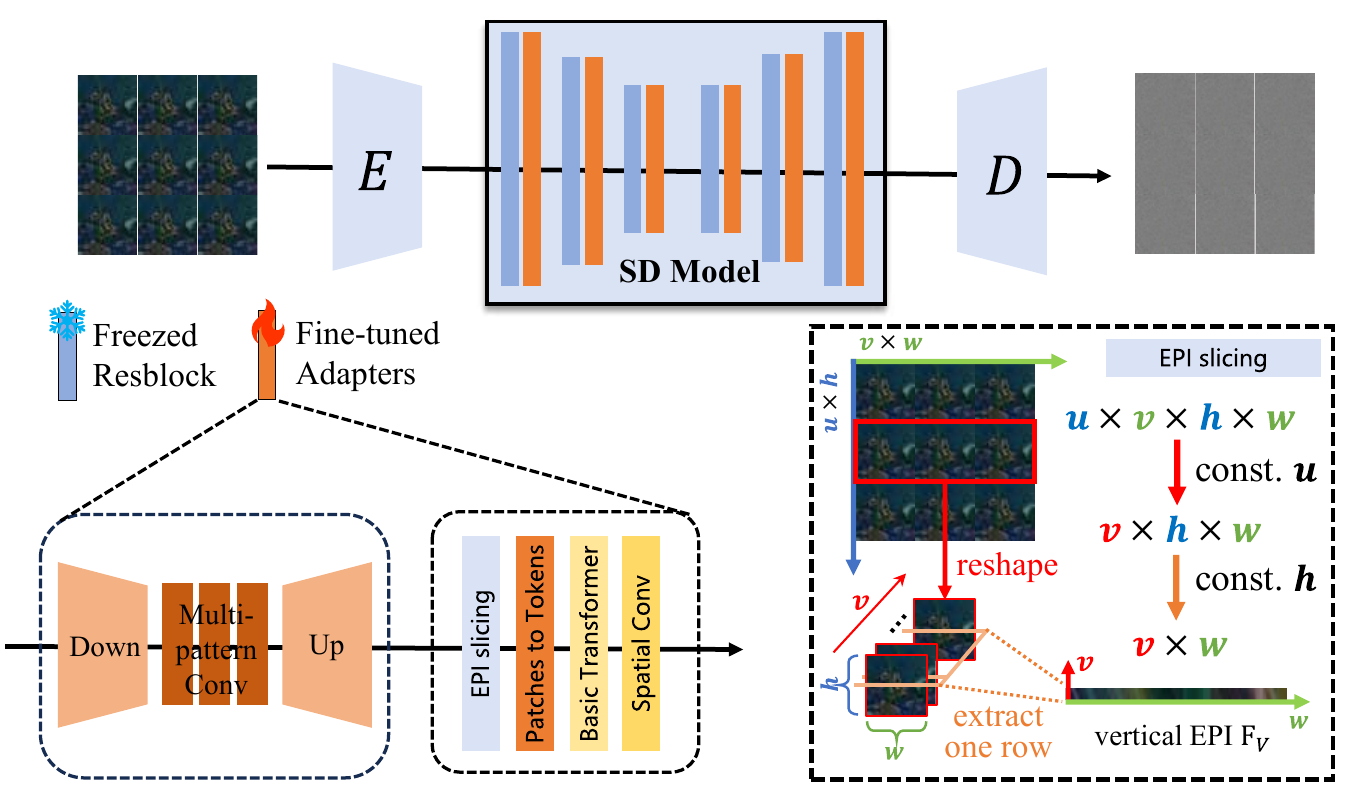}};
        \node[align=center, font=\small] at (2.4,-0.1) {\textbf{(a)} Conv-Adapter};
        \node[align=center, font=\small] at (6,-0.1) {\textbf{(b)} EPIT-Adapter};
        \node[align=center, font=\small] at (11.4,-0.1) {\textbf{(c)} EPI slicing};
    \end{tikzpicture}
    \caption{Architecture of our adapted SD-Turbo. Two types of adapters are introduced: a Conv-Adapter for capturing multi-dimensional features and an EPIT-Adapter for multi-view correlation. }
    \label{fig-adapters}
\end{figure*}

In the diffusion training process, as detailed in Algorithm~\ref{alg:forward}, the denoiser network $\epsilon_{\theta}$ is trained to estimate the added noise, enabling iterative recovery of the clean sample $\mathbf{X}_0$ from the fully noised state $\mathbf{X}_T$. As $t$ increases toward $T$, $\mathbf{X}_t$ becomes increasingly corrupted until $\mathbf{X}_T$ closely approximates a standard Gaussian distribution $\mathcal{N}(0, \mathbf{I})$. Consequently, $\epsilon_\theta$ learns to reconstruct meaningful $\mathbf{X}_0$ from pure Gaussian noise. The training objective simplifies to the following L2 loss:
\begin{equation}
    \left\| \epsilon-\epsilon_{\theta}(\mathbf{X}_t,\mathbf{Y}_0,t) \right\|_2^2.
\end{equation}

During inference, as shown in Algorithm~\ref{alg:reverse}, a valid image $\mathbf{X}_0$ is generated by starting from Gaussian noise $\mathbf{X}_T \sim \mathcal{N}(0, \mathbf{I})$ and iteratively denoising. Conventional diffusion models~\cite{ho2020denoising} require hundreds to thousands of steps to produce high-quality results, incurring substantial computational cost. Recent acceleration techniques, such as DDIM~\cite{song2020denoising} and other few-step methods~\cite{lin2024common,yue2025arbitrary}, enable high-quality generation with significantly fewer steps. Moreover, the distilled SD-Turbo model~\cite{Rombach_2022_CVPR} achieves comparable image synthesis quality in as few as $1\sim4$ steps, providing an efficient foundation for our framework.

\section{Proposed Method}
\label{sec:method}
\subsection{Motivation}
Recent advances in diffusion-based restoration approaches~\cite{yue2023resshift,hou2023global,yue2024difface} have demonstrated that incorporating global guidance into the denoising process can effectively regulate the optimization trajectory and substantially improve restoration performance.
In underwater imaging, geometric information plays a pivotal role, as geometric distortion is a key factor that exacerbates underwater degradation. 
Moreover, unlike traditional 2-D imaging, 4-D LF imaging jointly captures spatial and angular information, providing rich geometric cues that facilitate accurate depth estimation and physical transmission modeling. These geometric cues naturally serve as powerful global guidance for enhancing underwater imaging.

Motivated by these insights, we propose an adapted diffusion-based enhancement framework built upon SD-Turbo~\cite{Rombach_2022_CVPR,sauer2024adversarial}, specifically tailored for underwater 4-D LF imaging by explicitly exploiting its inherent geometric information as guidance. 
Our framework introduces three key components: 
(a) an efficient noise prediction strategy to improve restoration quality and computational efficiency,
(b) a modified network architecture equipped with adapters to efficiently handle 4-D LFs, and 
(c) a global geometry regularization that integrates geometric cues to guide the denoising process.
The overall framework of the proposed GeoDiff-LF is presented in Fig.~\ref{fig-pipeline}.

\begin{figure*}
    \centering
    \includegraphics[width=0.9\linewidth]{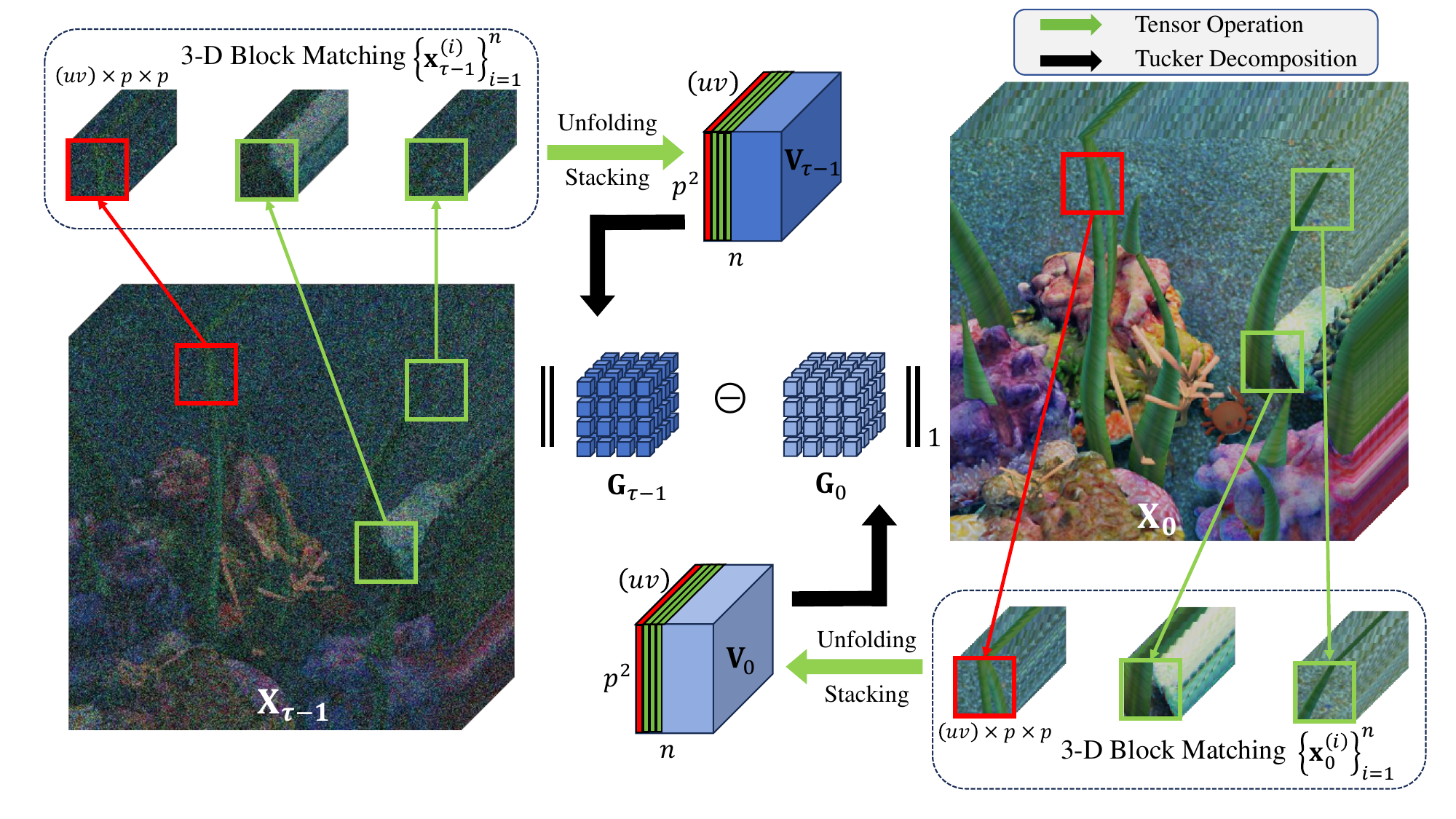}
    \caption{The entire process of global geometry regularization. We explore the intrinsic geometric structure of 4-D LFs from tensor-dimension similarity, which is measured by the core tensor of similar 3-D blocks from 4-D LFs.}
    \label{fig-tensor}
\end{figure*}

\subsection{Efficient Noise Prediction Based on Degradation}
\label{sec:noise}
The entire process of diffusion models involves long sampling intervals and high randomness in noise schedules, which significantly increases computational costs. 
Considering diffusion reversion in the context of underwater image enhancement, the underwater image $\mathbf{Y}_0$ only slightly deviates from the target clean image $\mathbf{X}_0$ in most cases, primarily in depth-dependent attenuation and scattering components. Therefore, inspired by~\cite{yue2025arbitrary}, we initiate the sampling process from an intermediate timestep $\tau~(\tau<T)$ with a SNR(signal-to-noise ratio) threshold of 1.38 in the widely used Stable Diffusion~\cite{sauer2024adversarial}, thereby preserving the stages of the diffusion model responsible for recovering underwater degradation.
Simultaneously, we introduce a noise map prediction network that predicts noise from the forward process to inform the reverse denoising process. 
As shown in Fig.~\ref{fig-pipeline}(a), the network with parameter $w$, denoted as $f_w$, takes the underwater images $\mathbf{Y}_0$ and the timestep $\tau$ as input and then outputs the desired noise maps $f_w(\mathbf{Y}_0,\tau)$, which depend on the degradation degree of the underwater images.

Because the underwater images $\mathbf{Y}_0$ and the high-quality images $\mathbf{X}_0$ become increasingly indistinguishable when perturbed by Gaussian noise with an appropriate magnitude, we choose an optimal noise map $f_{w}(\mathbf{Y}_0,\tau)$ to perturb $\mathbf{Y}_0$ and then the adapted diffusion model can generate the corresponding initial noisy sample $\mathbf{X}_{\tau}$ that is defined as follows:
\begin{equation}
\mathbf{X}_{\tau} = \sqrt{\bar{\alpha}_{\tau}}\mathbf{Y}_0 + \sqrt{1-\bar{\alpha}_{\tau}}f_{w}(\mathbf{Y}_0,\tau),~\tau<T.
\end{equation}

This intermediate step serves as an efficient starting point for sampling, reducing the number of denoising iterations required. By integrating this noise prediction mechanism into our entire framework, we can optimize sampling efficiency while maintaining high-quality restoration, making our training and testing processes computationally practical for addressing 4-D LF images.

\subsection{Geometry-Aware Diffusion Adapters}
Given the high computational cost of training a 4-D LF-based diffusion model from scratch, we adopt an efficient strategy by freezing a large-scale pre-trained diffusion model (e.g., SD-Turbo~\cite{Rombach_2022_CVPR,sauer2024adversarial}) and adapting it to the 4-D LF domain. To fully exploit the inherent geometric structure of 4-D LFs, we introduce several lightweight and learnable adaptation modules into both convolutional and attention blocks, as shown in Fig.~\ref{fig-adapters}. 
To facilitate understanding, we walk through the process from noisy input samples to predicted noise outputs. For the noisy LF samples $\mathbf{X}_{\tau}\in \mathbb{R}^{b\times u\times v\times h\times w\times c}$, where $b$ denotes batch size, $u, v$ denote angular dimensions, and $h, w$ denote spatial dimensions, we first treat all views as one batch and pass them collectively through the encoder to obtain latent representations. These representations are then processed through the frozen ResNet blocks. Subsequently, we introduce two types of adapters to modulate the 4-D LF features.

For the convolutional component in Fig.~\ref{fig-adapters}(a), we employ a bottleneck architecture comprising two fully connected layers and three multi-pattern convolutional layers~\cite{lin2025enhancing, lyu2024enhancing}. The multi-pattern convolutions incorporate operations in spatial, angular, vertical EPI, and horizontal EPI dimensions, where an EPI (epipolar plane image) of 4-D LF is obtained by fixing one angular index and one spatial index, as shown in Fig.~\ref{fig-adapters}(c). The two fully connected layers, both using 3-D convolutions with $1\times1$ kernels, perform dimension reduction and restoration, respectively. The multi-pattern convolution operates in a low-dimensional space to capture inter-viewpoint correlations of 4-D LFs, ensuring low computational complexity. Formally, for an input feature $\mathrm{F}_1\in\mathbb{R}^{buv\times h_d\times w_d\times c_d}$, the introduced convolution adapter could be formally written as:
\begin{equation}
    \texttt{Conv-Adapter}(\mathrm{F_1})=\mathrm{F_1}+W_{up}(\texttt{MP}(W_{down}(\mathrm{F_1}))),
    \label{equ:conv}
\end{equation}
where $W_{up}(\cdot)$ and $W_{down}(\cdot)$ denote $1\times 1$ 3-D convolution, which are equivalent to learnable matrices of sizes $c_d \times c_l$ and $c_l \times c_d~\left(c_l<c_d\right)$, respectively; $\texttt{MP}(\cdot)$ denotes multi-pattern convolutions, which require simultaneous reshaping of four spatial and angular coordinates while applying each convolution pattern to only two dimensions once time.

\begin{algorithm*}[t]
\caption{Training Process of GeoDiff-LF}
\label{alg:overall-training}
\textbf{Require:} Normal and underwater LF image pairs $(\mathbf{X}_0,\mathbf{Y}_0)$, diffusion timesteps $\mathcal{S}_0$, balancing parameter $\lambda$, noise map predictor $f_w$, and a pre-defined model $\epsilon_{\theta}$, $\theta=\{ \theta_0, \theta_1\}$, with freezed SD-Turbo parameters $\theta_0$ and initialized adapters parameters $\theta_1$.
\begin{algorithmic}[1]
\State Initialize parameters $w$ and $\theta_1$.
\State \textbf{Repeat}
\State ~~~~Sample an LF image pair $(\mathbf{X}_0,\mathbf{Y}_0)$, $\tau\sim\mathcal{S}_0$, $\bar{\alpha}_{\tau}\sim p(\bar{\alpha})$ and noise $\mathrm{Z}_{\tau-1} \sim \mathcal{N}(0,\bm{I})$
\State ~~~~$\mathbf{X}_{\tau} = \sqrt{\bar{\alpha}_{\tau}}\mathbf{Y}_0 + \sqrt{1-\bar{\alpha}_{\tau}}f_{w}(\mathbf{Y}_0,\tau)$
\State ~~~~Construct the intermediate reconstructed sample $\mathbf{X}_{\tau-1}$ via Eq.~\eqref{eq:inter-recon}
\State ~~~~Perform a gradient descent step on $\nabla_{w,\theta_1} \left\{ \left\| \mathbf{X}_0 - \hat{\mathbf{X}} \right\|_1 + \lambda \mathcal{L}_s^{\tau}\right\}$ 
\State \textbf{Until} converged
\State \textbf{Return} the resulting diffusion model $\epsilon_{\tilde{\theta}}$ and noise predictor $f_{\tilde{w}}$
\end{algorithmic}
\end{algorithm*}

The original attention block of the U-Net only performs self-attention for single 2-D images, neglecting the information of the angular dimension, while joint-spatial-angular attention, as demonstrated in~\cite{liang2023learning, chao2024bigepit}, can effectively model angular dependencies of 4-D LFs.
We incorporate the EPIT block~\cite{liang2023learning} into the attention block to enable modeling of correlations in the angular dimension of 4-D LFs, as shown in Fig.~\ref{fig-adapters}(b). EPIT-Adapters initially separately reshape the feature $\mathrm{F}_2 \in \mathbb{R}^{buv\times h_d\times w_d\times c_d}$ into the horizontal EPI pattern $\mathrm{F}_{H} \in \mathbb{R}^{buh_d\times v\times w_d\times c_d}$ and the vertical EPI pattern $\mathrm{F}_{V} \in \mathbb{R}^{bvw_d\times u\times h_d\times c_d}$,
one of which is shown in Fig.~\ref{fig-adapters}(c). These partitioned features are subsequently converted to the sequences of tokens along non-constant dimensions. For example, the horizontal EPI features $\mathrm{F}_{H} \in \mathbb{R}^{uh_d\times bvw_d\times c_d}$ are passed through a linear projection matrix $W_{in} \in \mathbb{R}^{c_d\times c_k}$, where $c_k$ denotes the embedding dimension of each token. Then, they are passed through a basic transformer for self-similarity calculation and are finally converted to a spatial convolution layer. 

The Conv-Adapter employs multi-pattern convolutions that operate separately on spatial and angular dimensions. In underwater scenes, depth-dependent color attenuation and scattering manifest differently across these dimensions: the spatial domain suffers from non-uniform color casts and contrast loss, while the angular domain exhibits view-dependent scattering noise. By disentangling spatial and angular processing, the Conv-Adapter can simultaneously correct local color deviations in each view and suppress view-consistent scattering artifacts through angular filtering. The EPIT-Adapter explicitly models cross-view correlations through EPIs. Underwater scattering introduces floating particles, breaking the ideal straight-line EPI structure. The EPIT-Adapter learns to re-establish robust angular correspondences despite these perturbations. This is fundamentally different from in-air LF processing, where EPI lines are clean, and the main challenge is disparity estimation; underwater, the primary difficulty is recovering corrupted EPI structure from severe scattering. Thus, their application to the coupled spatial-angular degradation of underwater imaging is task-specific. This dual-adapter architecture facilitates robust geometric modeling, enabling the denoising network to handle the intricate spatial and angular relationships inherent in underwater LF images. More importantly, these adapters serve as the geometric interface that translates the 2-D diffusion prior into the 4-D LF domain without destroying the original generative capacity.

\subsection{Global Geometry Regularization}

The diffusion reverse process is inherently stochastic; without explicit guidance, the random noise schedule disrupts the delicate 4-D geometric structure. 
Unlike the Adapters that improve LF geometric information extraction through network architecture design, we propose a regularization to further exploit the geometric information of 4-D LFs and thereby optimize the denoising trajectory. 
During the reverse phase of diffusion models, the learnable denoiser $\epsilon_\theta (\mathbf{X}_{\tau},\mathbf{Y}_0,\tau)$ estimates the corresponding noise from the input noisy sample $\mathbf{X}_\tau$ at the pre-defined timestep $\tau$. Based on this noise, we can obtain the intermediate reconstructed sample $\mathbf{X}_{\tau-1}$, formulated as Eq.~\eqref{eq:inter-recon}. Consequently, during training, we can apply additional regularization to the intermediate reconstruction $\mathbf{X}_{\tau-1}$, steering the diffusion toward the intrinsic low-rank manifold of clean LF tensors. Note that one may instead regularize the noisy sample $\mathbf{X}_{\tau}$, but the performance could be dropped as discussed in Section~\ref{sec:ablation}.

To guide the network in learning global geometric structure representations, we treat a group of 4-D LF images $\mathbf{X}_{0} \in \mathbb{R}^{uv\times h\times w\times 3}$ as a multi-dimensional tensor and leverage its intrinsic structure for regularization, which is shown as Fig.~\ref{fig-tensor}. We first divide it into a set of 3-D non-overlapping blocks $\left\{\mathbf{x}_{0}^{(i)}\right\}_{i=1}^{n}$, where $\mathbf{x}_{0}^{(i)} \in \mathbb{R}^{uv\times p^2 \times 3}$ denotes the matrixization of a small block from $\mathbf{X}_{0}$, where $p$ is the patch size of small block on spatial dimension.
We then use a 3-D block-matching approach to group these blocks into $k$ similar clusters. 
By aggregating the matrices within each cluster into a compact tensor, denoted as $\left\{\mathbf{V}_0^{n_j}\right\}_{j=1}^k$, the core tensor obtained through tensor Tucker decomposition~\cite{kolda2009tensor} effectively characterizes the global geometric structure in a low-rank framework. At last, we then apply the same methodology to intermediate reconstruction $\mathbf{X}_{\tau-1}$, resulting in  $\left\{\mathbf{V}^{n_j}_{\tau-1}\right\}_{j=1}^k$, and regularize
\begin{equation}
\begin{aligned}
&\mathcal{L}_{s}(\mathbf{V}^{n_j}_{\tau-1},\mathbf{V}_{0}^{n_j}) = \left\| \textbf{G}_{\tau-1}^j - \textbf{G}_{0}^j \right\|_1, 
~j=1,2,\dots,k,\\
&\mathbf{V}_{t}^{n_j} = \textbf{G}_t^j \times_1 \textbf{O}_t^j \times_2 \textbf{P}_t^j \times_3 \textbf{Q}_t^j,~t=0,1,\dots,\tau-1,
\end{aligned}
\end{equation}
where $\textbf{G}_t^j \in \mathbb{R}^{r_1 \times r_2 \times r_3}$ is the core tensor, and $ \textbf{O}_t^j \in \mathbb{R}^{I_1 \times r_1}$, $ \textbf{P}_t^j \in \mathbb{R}^{I_2 \times r_2} $,  $\textbf{Q}_t^j \in \mathbb{R}^{I_3 \times r_3} $ are the factor matrices of $\mathbf{X}_t$'s tensor Tucker decomposition. 
Such a global regularization enforces consistency in the tensor structure, which effectively reduces the dimensionality of the feature space and yields a more compact representation.

\begin{table*}[t]
\scriptsize
  \caption{Quantitative comparisons (PSNR/SSIM/LPIPS/$\vartriangle$$E$) of different methods on the LFUB dataset. The best second-best results are highlighted in \textbf{bold} and \underline{underline}, respectively. ``↑'' (resp. ``↓'') means the larger (resp. smaller), the better. All compared methods are retrained on the LFUB training set and evaluated on the testing set.}
  \label{tab-results}
  \centering
  \small
  \setlength{\tabcolsep}{2mm}{
  \begin{tabular}{c|ccccccc}
    \toprule[1.2pt]
    ~~~Methods~~~ 
    &~Input~
    &~Fusion~\cite{ancuti2012enhancing}~
    &~GDCP~\cite{peng2018generalization}~
    &~WWPF~\cite{zhang2023underwater}~
    &~LANet~\cite{liu2022adaptive}~ 
    &~PUIE~\cite{fu2022uncertainty}~ 
    &~Ushape~\cite{peng2023u}~ \\
    
    \midrule
    PSNR $\uparrow$    
    &11.22&14.19 &13.23 &14.67 &17.97 &18.81  &18.29 \\
    SSIM $\uparrow$    
    &0.2056&0.3470 &0.4118  &0.5304 &0.7987 &0.7768  &0.7877       \\
    LPIPS $\downarrow$    
    &0.7394&0.6512 &0.6851  &0.5408 &0.3091 &0.2918  &0.3157       \\    
    $\vartriangle$$E$ $\downarrow$   
    &25.05&24.35 &26.83 &19.25 &16.43 &14.37  &15.88      \\   
    
    \midrule[1.2pt]
     ~~~Methods~~~  
    &~Uranker~\cite{guo2023underwater}~
    &~UVE~\cite{du2024end}~
    &~UIEDP~\cite{du2025uiedp}~
    &~DistgSSR~\cite{wang2022disentangling}~
    &~MSPNet~\cite{wang2023multi}~
    &~LFUB~\cite{lin2025enhancing}~
    &~Ours~ \\
    \midrule
    PSNR $\uparrow$    
     &18.48  &19.91&21.95 &17.52 &18.92 &\underline{22.51}  &\textbf{23.67}   \\
    SSIM $\uparrow$    
     &0.8345 &0.8571&0.8658 &0.7631 &0.8226 &\underline{0.8680}  &\textbf{0.8711}    \\
    LPIPS $\downarrow$    
    &0.2874 &0.2733&0.2657  &0.3110 &0.3269 &\underline{0.2535}  &\textbf{0.2361}      \\    
    $\vartriangle$$E$ $\downarrow$   
    &14.00 &14.42&13.84 &16.58 &15.43 &\underline{13.51}  &\textbf{11.71}       \\   
    \bottomrule[1.2pt]
  \end{tabular}}
\end{table*}

Throughout the training process, applying the regularization above presents challenges: first, the influence of global geometry regularization is weaker at earlier time steps; second, its initial application may reduce sample diversity and quality. To address these issues, we propose a progressive and adaptive regularization strategy. The improved regularization schedule is formulated as:
\begin{equation}
\begin{aligned}
    &\mathcal{L}_s^t = \rho_t \mathcal{L}_{s}\left(\mathbf{V}_{t-1}^{n_j},\mathbf{V}_{0}^{n_j}\right),~
    j=1,2,\cdots,k,\\
    &\rho_t \sim \left\{\bar{\alpha}_1^2, \bar{\alpha}_2^2,...,\bar{\alpha}_{\tau}^2\right\},~t = 2,3,...,\tau,
\end{aligned}
\label{eq:Weight_reg}
\end{equation}
where $\rho_t$ denotes an adaptive factor for regularization function $\mathcal{L}_{s}\left(\mathbf{V}_{t-1}^{n_j},\mathbf{V}_{0}^{n_j}\right)$. This strategy enables the model to adaptively handle diverse samples across different timesteps, gradually regularizing intermediate reconstructions while effectively suppressing noise through an aligned schedule.

\textbf{Remark:} \emph{Before proceeding, we discuss more about the advantages of the proposed tensor-based global geometry regularization.}

\emph{First, one may be concerned about why not directly apply EPI-based constraints, which are widely adopted for local geometry preservation. The main reason is that they inherently operate on 2-D slices by fixing one spatial and one angular dimension. Consequently, such losses can only enforce local epipolar straightness and lack the capacity to regularize the joint spatial-angular structure across the full 4-D volume. Therefore, we do not employ an EPI-based loss for global regularization. Instead, the EPIT-Adapter is responsible for extracting local EPI features and angular correlations, while the proposed Tucker-core regularization operates on 3-D block-matched tensors to capture global low-rank correlations across all spatial and angular modes. These two designs ensure that local epipolar geometry and holistic volumetric consistency are jointly preserved in a complementary manner.}

\emph{Second, one may also be concerned about why not use pixel-level losses, which could be easier to implement and more efficient. However, the underwater degradation severely corrupts the intrinsic structure of LF images. In this scenario, pixel-level losses merely impose an average penalty per pixel without exploiting the underlying geometric structure; as a result, the reconstruction suffers from high variance and fails to reliably recover the lost spatial-angular dependencies. By contrast, the low-rank Tucker-core constraint explicitly regularizes the global feature subspace, suppressing inconsistent fluctuations and enabling more stable restoration of the LF structure. Empirically, adding the global geometry regularization increases the training time by only 0.1 extra seconds per iteration (from 0.234 seconds to 0.333 seconds). Therefore, our regularization remains computationally acceptable compared to pixel-level losses.}

\subsection{Model Training and Inference}
To train the whole pipeline, we adopt a pixel-level objective, and our global geometry regularization, and the overall loss function is expressed as:
\begin{equation}
    \left\| \mathbf{X}_0 - \hat{\mathbf{X}} \right\|_1 + \lambda \mathcal{L}_s^t,
    \label{eq:loss}
\end{equation}
where $\hat{\mathbf{X}}$ is the denoised result from $\mathbf{X}_{\tau}$, as follows:
\begin{equation}
    \hat{\mathbf{X}} = \frac{1}{\sqrt{\bar{\alpha}_{\tau}}}\left( \mathbf{X}_{\tau} - 
    \sqrt{1-\bar{\alpha}_{\tau}}\epsilon_{\theta}(\mathbf{X}_{\tau},\mathbf{Y}_0,\tau)\right),
\end{equation}
and $\lambda$ denotes the balancing parameter, set to 1. Based on the discussion in Section~\ref{sec:noise}, we set the timestep $\tau \leq 500$ and randomly select a starting timestep from the diffusion timesteps $\mathcal{S}_0=\left\{500,400,300,200\right\}$, inspired by the skipping strategy~\cite{lin2024common}. Further details of the training process are provided in Algorithm~\ref{alg:overall-training}. Notably, the overall loss in Eq.~\eqref{eq:loss} is optimized with respect to both the noise map predictor parameters $\mathbf{w}$ and the adapter parameters $\theta_1$, while the pre-trained SD-Turbo parameters $\theta_0$ remain frozen. The gradients backpropagate through the entire reverse diffusion trajectory from the final reconstruction $\hat{\mathbf{X}}$ to both $\mathbf{w}$ and $\theta_1$. We emphasize that no explicit ground-truth supervision is imposed on the predicted noise map $f_{\mathbf{w}}(\mathbf{Y}_0, \tau)$, because the forward noise $\boldsymbol{\epsilon}$ is randomly sampled from a Gaussian distribution and thus lacks a deterministic target for each training sample. Instead, $f_{\mathbf{w}}$ is trained in a fully end-to-end manner: it learns a task-specific noise initialization conditioned on the underwater image $\mathbf{Y}_0$ and timestep $\tau$, with its optimization driven entirely by the pixel-level reconstruction loss and the global geometry regularization. This indirect supervision is stable because the lightweight architecture of $f_{\mathbf{w}}$ prevents overfitting, and the training signal from the final enhanced quality naturally guides the predictor toward producing noise maps that facilitate efficient few-step restoration.

During inference, five reversion steps, i.e., $\mathcal{S}=\left\{500,400,300,200,100\right\}$, are used throughout our experiments. The reverse process is thus applied exclusively at these key steps for each input. This strategy proves highly effective for two reasons: (1) the SD-Turbo model is inherently a distilled, few-step generative model designed for efficient sampling; (2) during training, we integrate a dedicated noise predictor and train the denoiser exclusively at these critical timesteps. Consequently, the model acquires robust cross-step denoising, enabling high-quality enhancement with very few sampling steps. The whole framework of inference is detailed in Algorithm~\ref{alg:overall-inference}.

\begin{algorithm}[H]
\caption{Inference Process of GeoDiff-LF}
\label{alg:overall-inference}
\textbf{Require:} Underwater LF images $\mathbf{Y}_0$, reversion timesteps $\mathcal{S}=\{\tau_i\}_{i=1}^s$, noise map predictor $f_{\tilde{w}}$, and diffusion model $\epsilon_{\tilde{\theta}}$.
\begin{algorithmic}[1]
\State $\mathbf{X}_{\tau_s} = \sqrt{\bar{\alpha}_{\tau_s}}\mathbf{Y}_0 + \sqrt{1-\bar{\alpha}_{\tau_s}}f_{\tilde{w}}(\mathbf{Y}_0,\tau_s)$
\State \textbf{for} $i=s,\dots,1$ \textbf{do}
\State ~~~~$\mathbf{Z}_{\tau_i}\sim\mathcal{N}(0,\bm{I})$ if $\tau_i >1$ else $\mathbf{Z}_{\tau_i}=0$
\State ~~~~$\mathbf{X}_{\tau_{i-1}}=g_{\tilde{\theta}}(\mathbf{X}_{\tau_i},\mathbf{Y}_0,\tau_{i-1}) + \sigma_{\tau_i}\mathbf{Z}_{\tau_i}$, where $g_{\tilde{\theta}}$ is defined in Eq.~\eqref{eq:g-theta}
\State \textbf{end for}
\State \textbf{Return} the enhancement results $\mathbf{X}_0$
\end{algorithmic}
\end{algorithm}

\section{Experiments}
\label{sec:exp}
In this section, we begin with an introduction to the experimental settings. Next, we compare our method against state-of-the-art methods and conduct a series of ablation studies to evaluate our framework. Finally, we show the difficult case to discuss the limitation.

\subsection{Experiment Settings}
\subsubsection{Training Details} We used the LFUB dataset~\cite{lin2025enhancing}, which includes a training set of 60 scenes and a test set of 15 scenes. The LF images feature an angular resolution of 7 $\times$~7 and a spatial resolution of 640 $\times$~360. During training, LF images were randomly cropped into fixed patches of 128 $\times$~128. The pixel values of all images were normalized to the range $[-1,~1]$. To augment the data, we applied random cropping, rotation, and flipping. 
Our framework was implemented in PyTorch using Python 3.9.0 and trained on a single NVIDIA RTX 3090 GPU with 24 GB of memory. We employed the Adam optimizer with $\beta_1 = 0.9$ and $\beta_2 = 0.999$. The training process involved over 2 million iterations with a batch size of 1 and a fixed learning rate of $2 \times 10^{-4}$. The noise map predictor $f_w$ is based on an encoder architecture comprising multi-pattern convolutional layers with a total parameter count of 30M, which is much lower than that of the frozen SD-Turbo backbone (857M parameters). It takes the underwater images $\mathbf{Y}_0$ and timestep $\tau$ as input, and outputs the predicted noise maps $f_w(\mathbf{Y}_0, \tau)$. This compact architecture inherently limits model capacity and reduces the risk of memorizing training-set-specific noise patterns. Moreover, 60 diverse underwater scenes with extensive data augmentation ensure that $f_w$ generalizes to varied degradation patterns rather than overfitting to particular image content.

\begin{figure*}[t]
\centering
\includegraphics[width=1\linewidth]{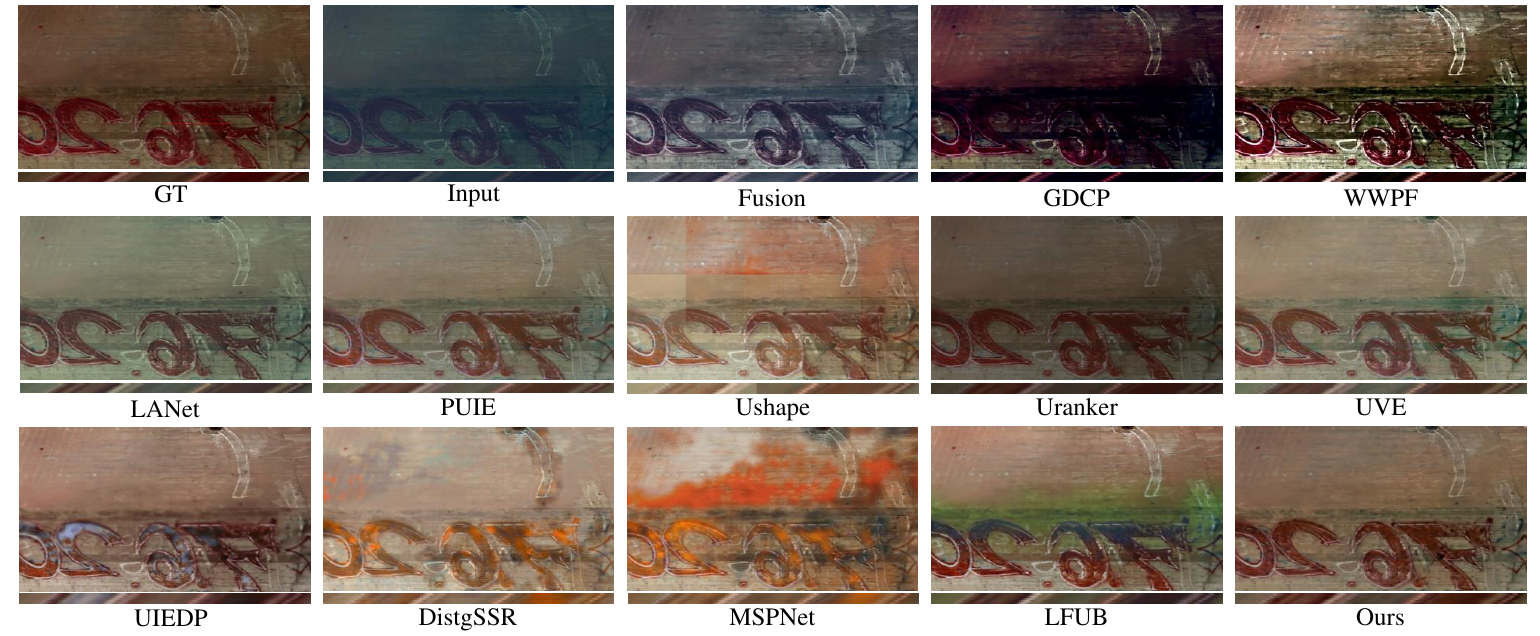}
\caption{Visual comparisons of the enhanced center SAI and EPI slice by different methods on the scene \textit{Boat} in the LFUB dataset. It is recommended to view this figure by zooming in.}
\label{fig:result-1}
\vspace{-3mm}
\end{figure*}

\begin{figure*}[t]
\centering
\includegraphics[width=1\linewidth]{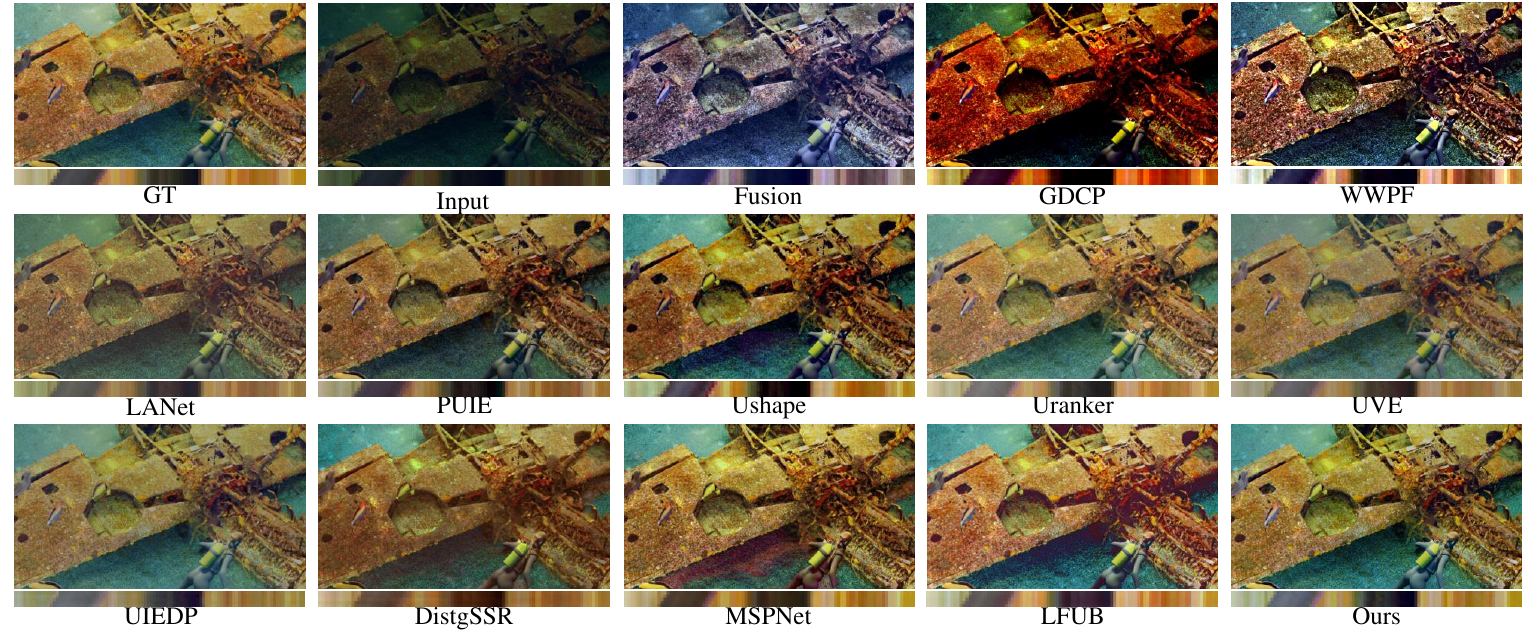}
\caption{Visual comparisons of the enhanced center SAI and EPI slice by different methods on the scene \textit{Airplane} in the LFUB dataset. It is recommended to view this figure by zooming in.}
\label{fig:result-2}
\vspace{-3mm}
\end{figure*}

\begin{table}
\caption{Qualitative comparisons (non-reference quality metrics UIQM/BRISQUE/NIMA/CCF) of different methods on the LFUID dataset. The best is highlighted in \textbf{bold}. ``↑'' (resp. ``↓'') means the larger (resp. smaller), the better.}
\label{tab-exp-real}
\centering
\small
\setlength{\tabcolsep}{2mm}{
\begin{tabular}{c|ccccc}
\toprule[1.2pt]
Method & UIQM $\uparrow$  & BRISQUE $\downarrow$ & NIMA $\uparrow$  & CCF $\uparrow$ \\ 
\midrule
WWPF~\cite{zhang2023underwater} & 4.499 & 38.615  & 3.098 & 31.186 \\
Ushape~\cite{peng2023u} & 3.653 & 37.927  & 3.245 & 15.771 \\
Uranker~\cite{guo2023underwater} & 4.199 & 36.459  & 3.461 & 20.285 \\
LFUB~\cite{lin2025enhancing} & 5.353 & 26.539  & 3.995 & 25.273 \\
LFUID~\cite{zhou2025lfuid} & 5.797 & 22.601  & 4.295 & \textbf{32.316} \\
Ours  & \textbf{6.038} & \textbf{20.871} & \textbf{4.692} & 29.221    \\
\bottomrule[1.2pt]
\end{tabular}}
\vspace{-3mm}
\end{table}

\begin{figure*}
    \centering
    \includegraphics[width=0.16\linewidth]{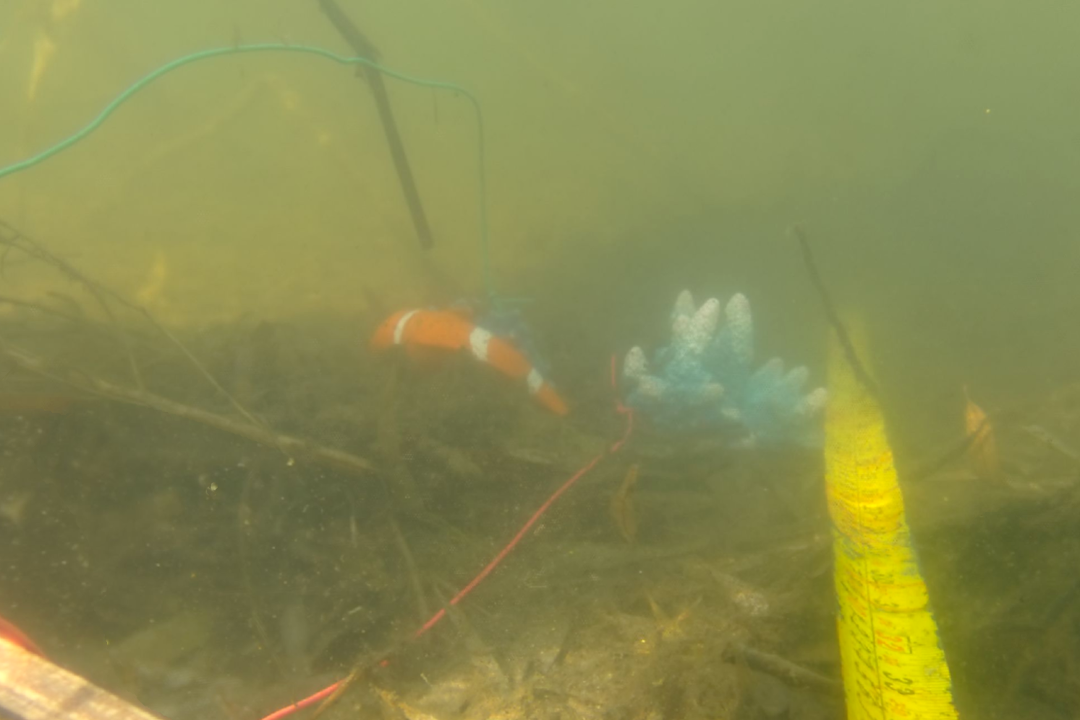}\hfill
    \includegraphics[width=0.16\linewidth]{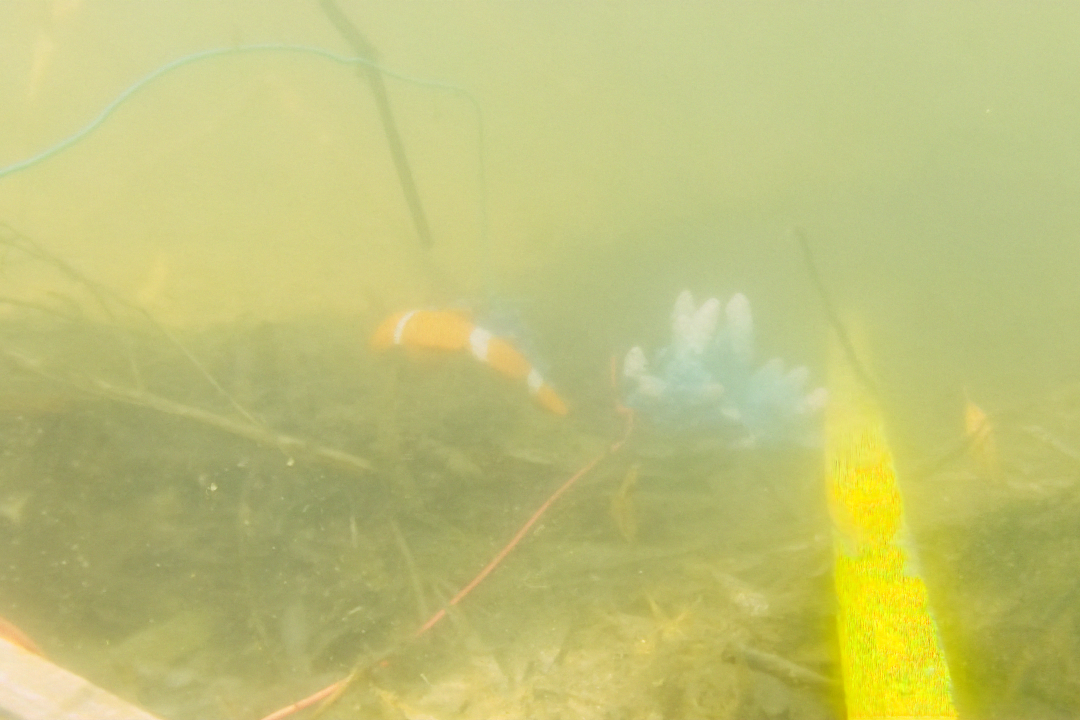}\hfill
    \includegraphics[width=0.16\linewidth]{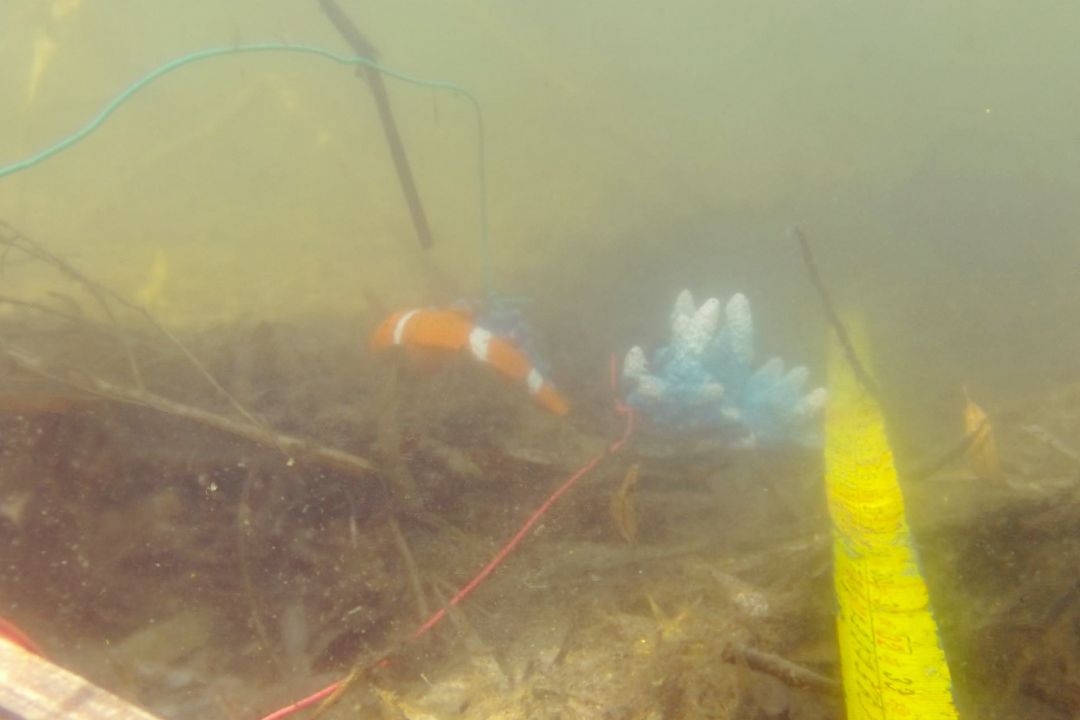}\hfill
    \includegraphics[width=0.16\linewidth]{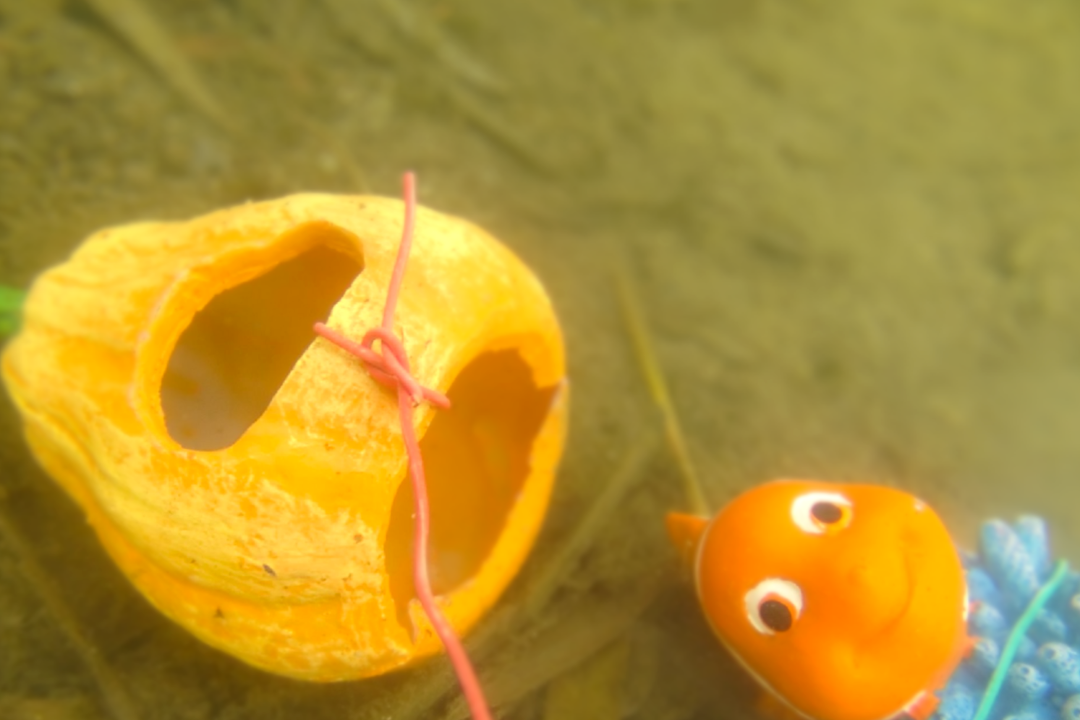}\hfill
    \includegraphics[width=0.16\linewidth]{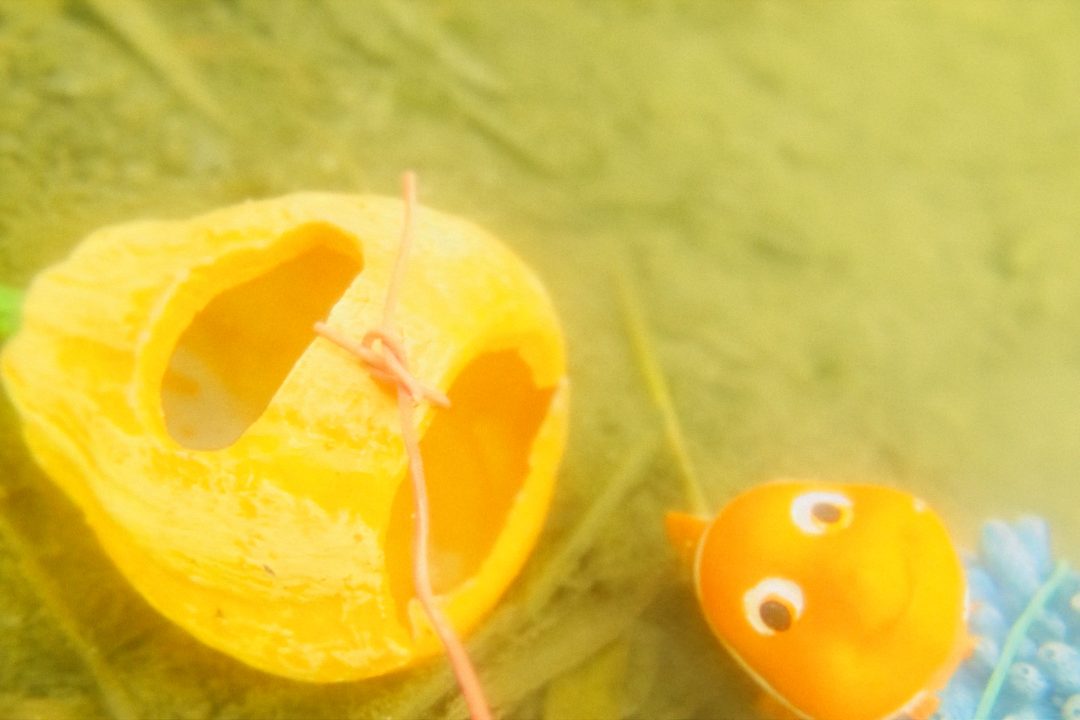}\hfill
    \includegraphics[width=0.16\linewidth]{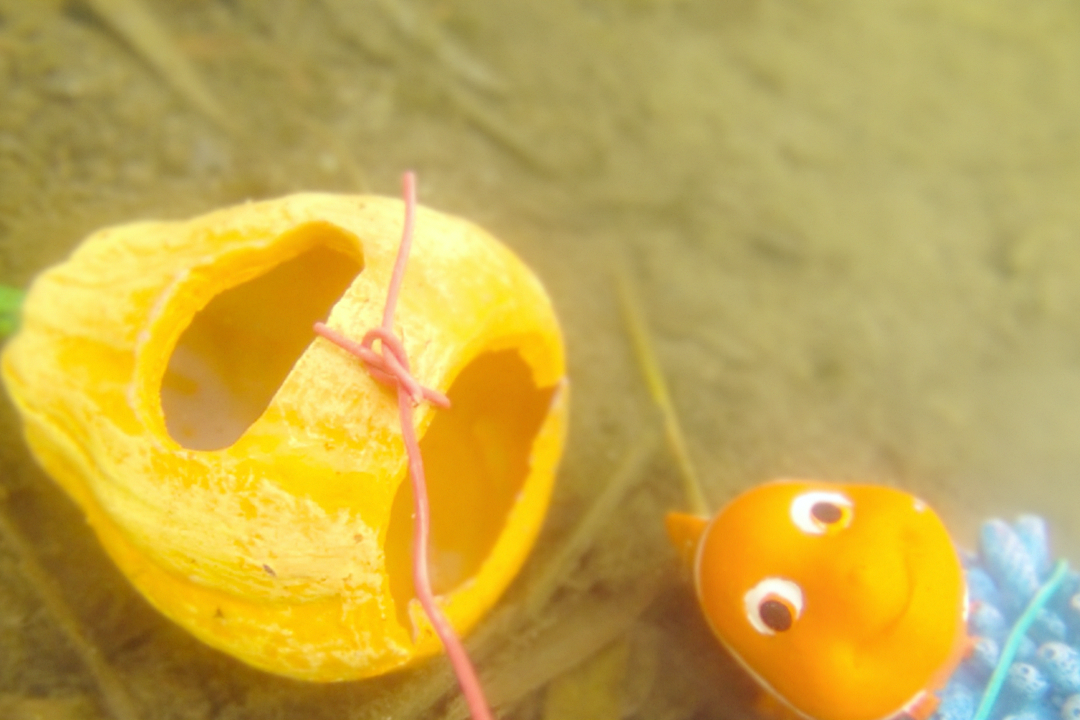} \\[4pt]
    \includegraphics[width=0.16\linewidth]{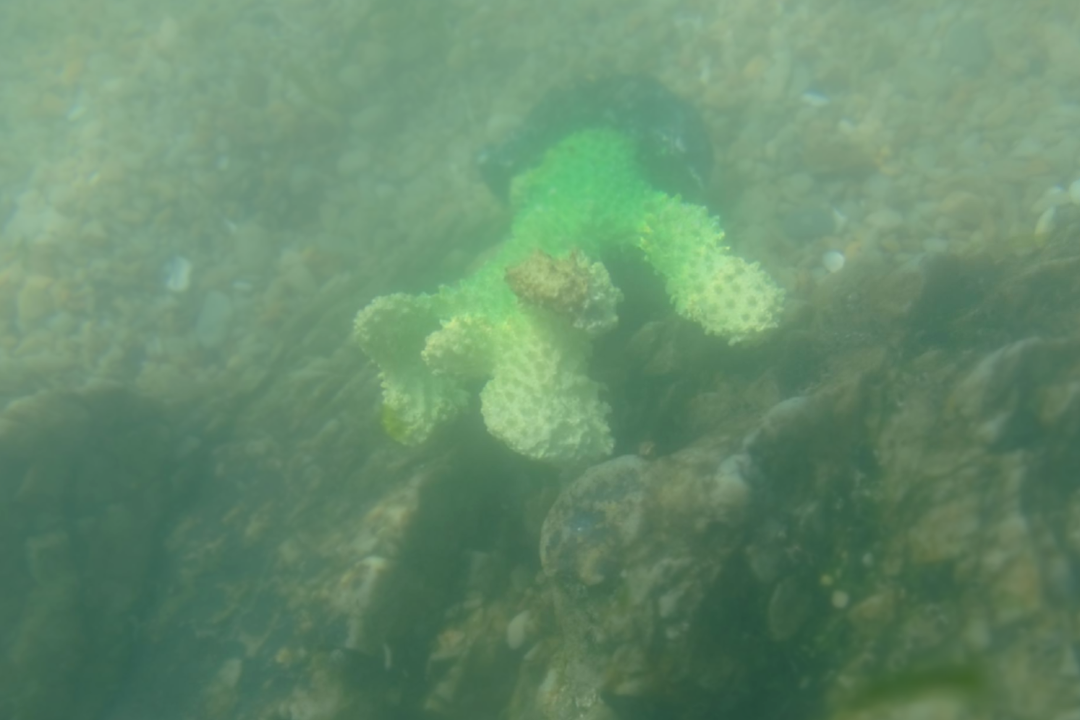}\hfill
    \includegraphics[width=0.16\linewidth]{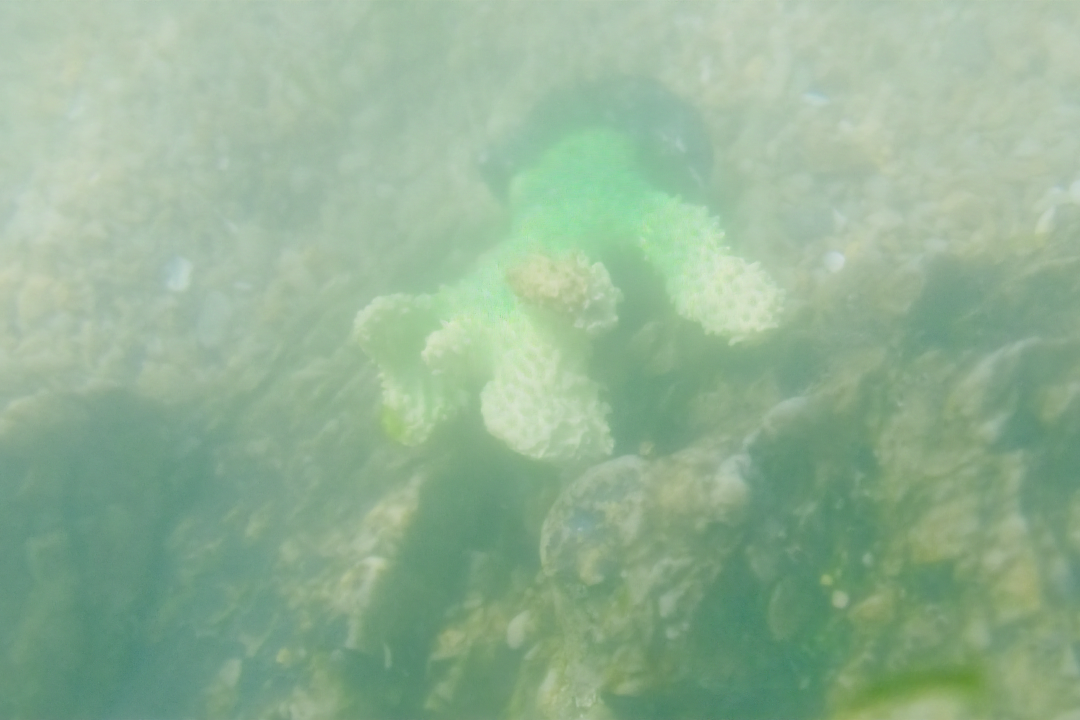}\hfill
    \includegraphics[width=0.16\linewidth]{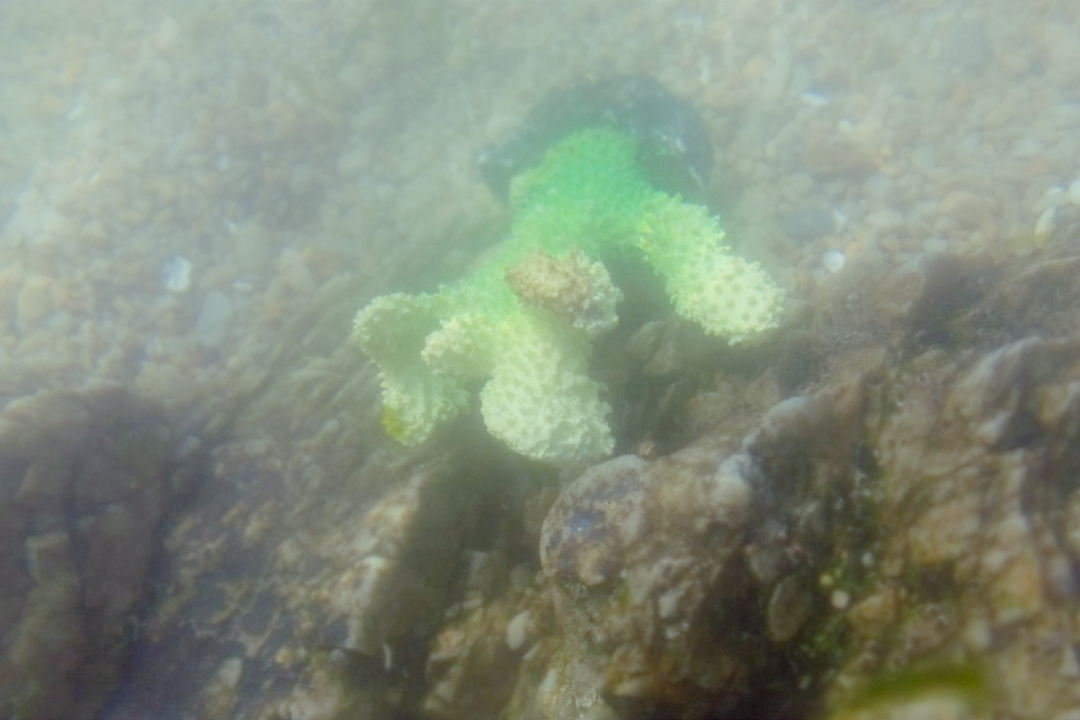}\hfill  
    \includegraphics[width=0.16\linewidth]{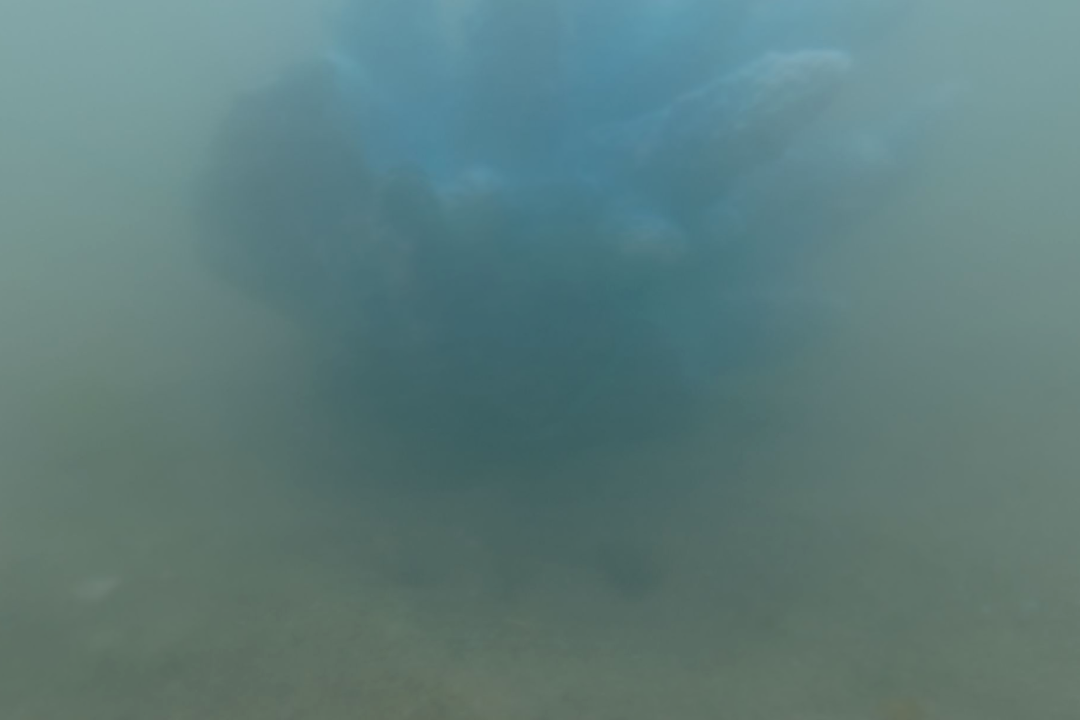}\hfill
    \includegraphics[width=0.16\linewidth]{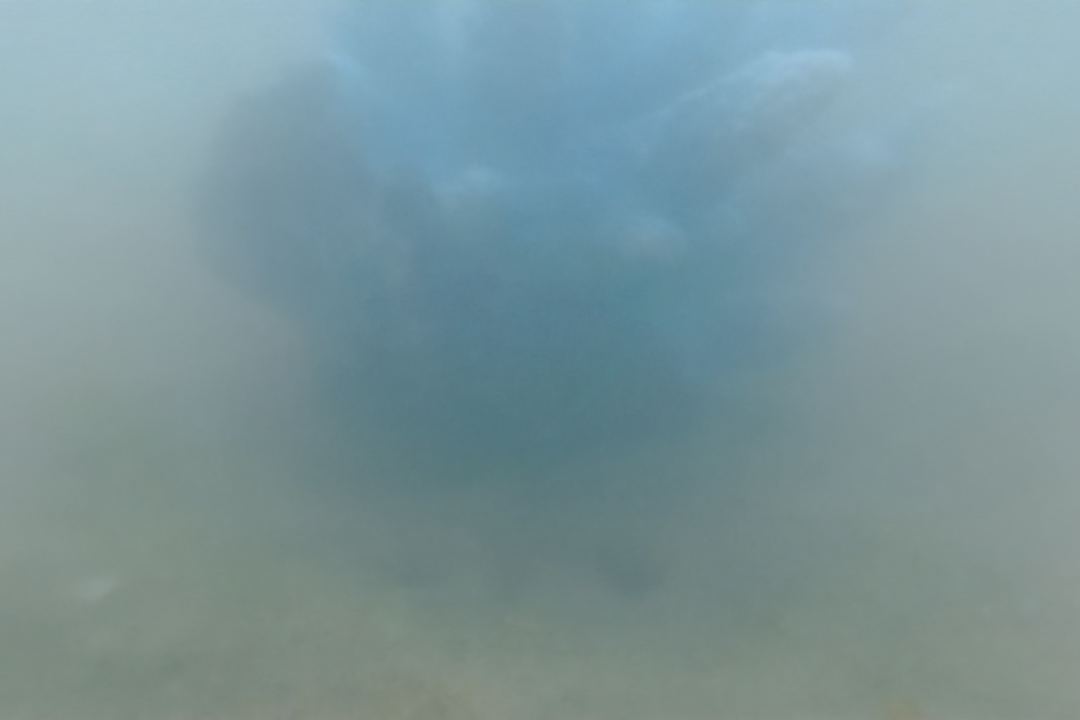}\hfill
    \includegraphics[width=0.16\linewidth]{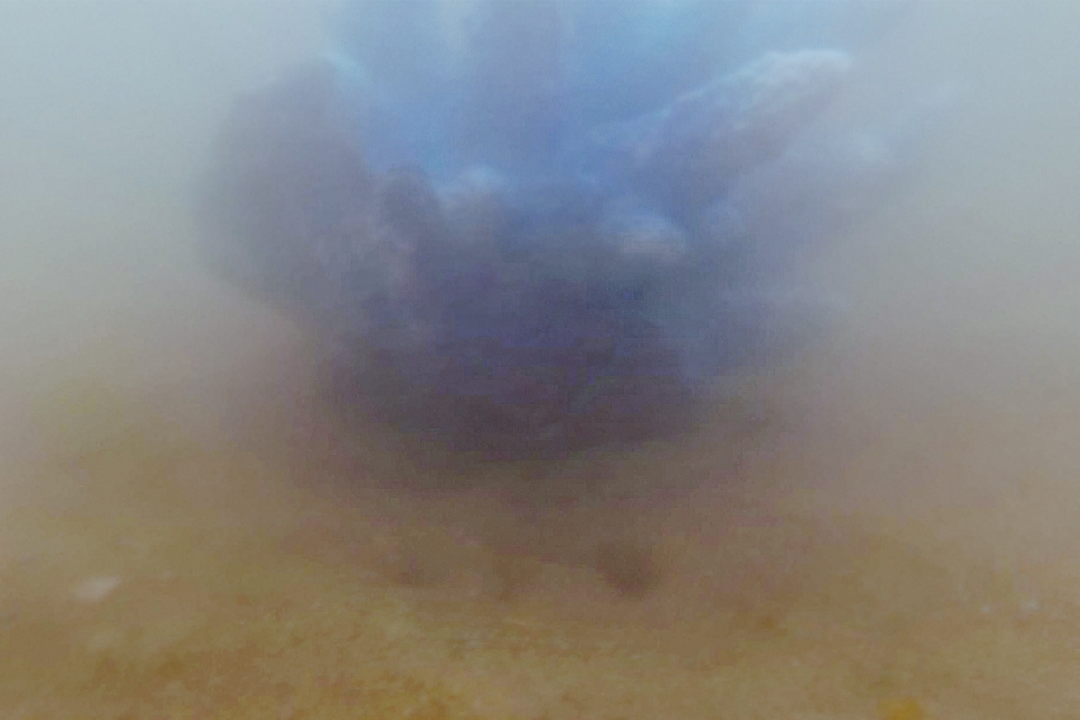}
    
    \small
    \makebox[0.16\linewidth]{(a) Input} \hfill
    \makebox[0.16\linewidth]{(b) LFUB~\cite{lin2025enhancing}} \hfill
    \makebox[0.16\linewidth]{(c) Ours} \hfill
    \makebox[0.16\linewidth]{(d) Input} \hfill
    \makebox[0.16\linewidth]{(e) LFUB~\cite{lin2025enhancing}} \hfill
    \makebox[0.16\linewidth]{(f) Ours}
    \caption{Visual comparisons of the enhanced center SAI by two underwater LF-based methods on the real captured dataset LFUID. It is recommended to view this figure by zooming in.}
    \label{fig:result-real}
    \vspace{-3mm}
\end{figure*}

\begin{figure*}
    \centering
    \includegraphics[width=0.16\linewidth]{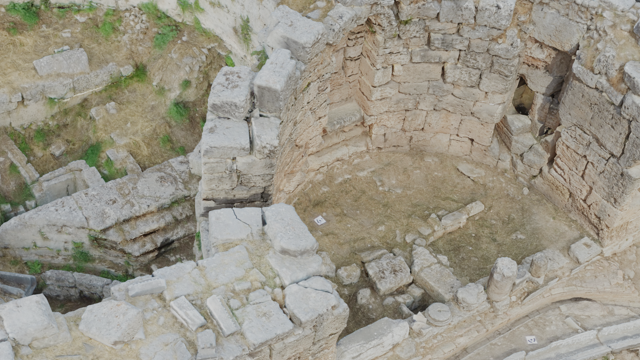}\hfill
    \includegraphics[width=0.16\linewidth]{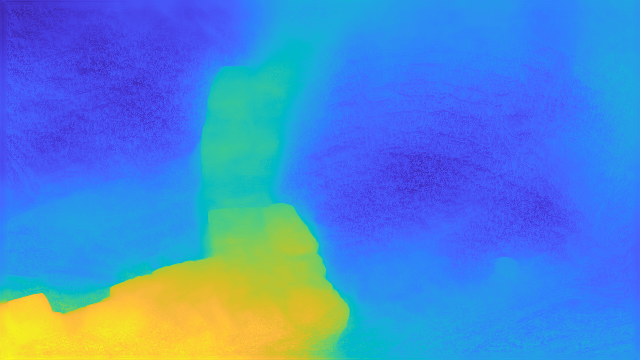}\hfill
    \includegraphics[width=0.16\linewidth]{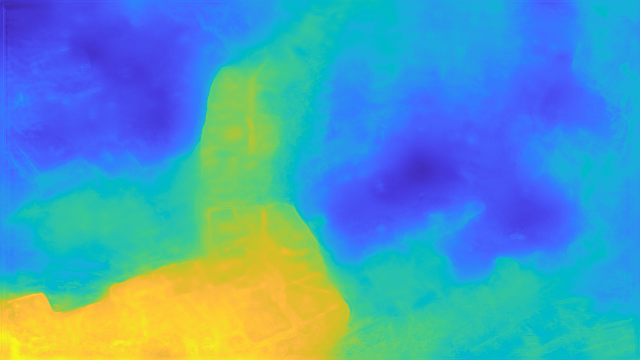}\hfill
    \includegraphics[width=0.16\linewidth]{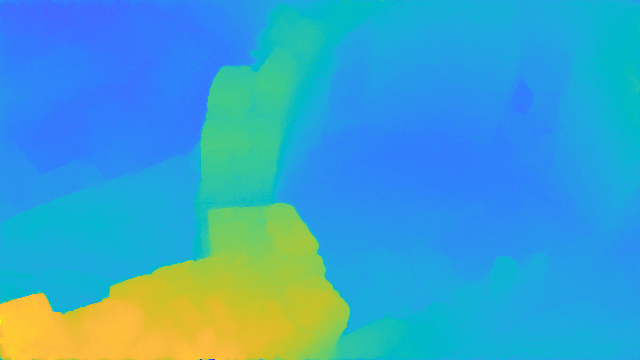}\hfill
    \includegraphics[width=0.16\linewidth]{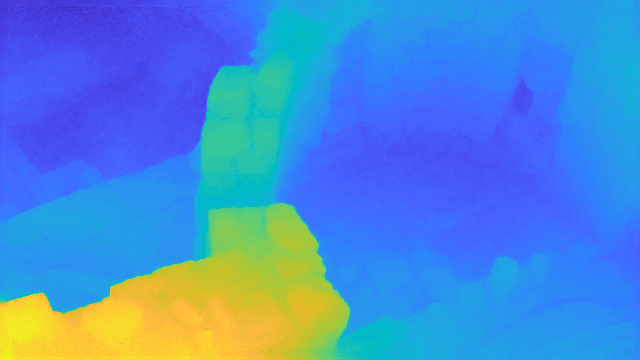}\hfill
    \includegraphics[width=0.16\linewidth]{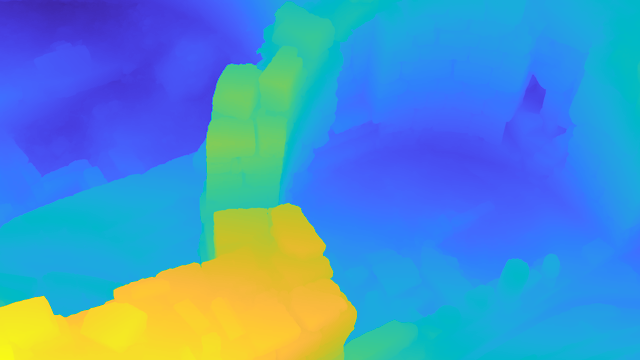} \\[4pt]
    \includegraphics[width=0.16\linewidth]{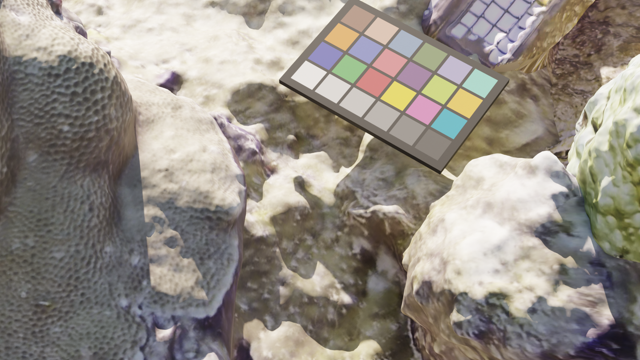}\hfill
    \includegraphics[width=0.16\linewidth]{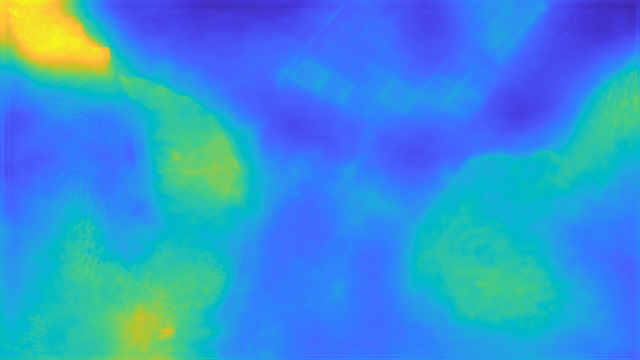}\hfill
    \includegraphics[width=0.16\linewidth]{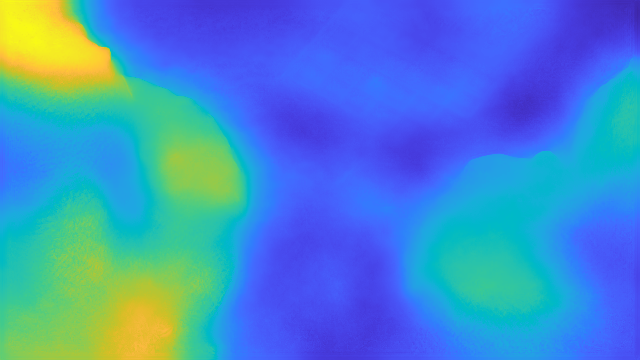}\hfill
    \includegraphics[width=0.16\linewidth]{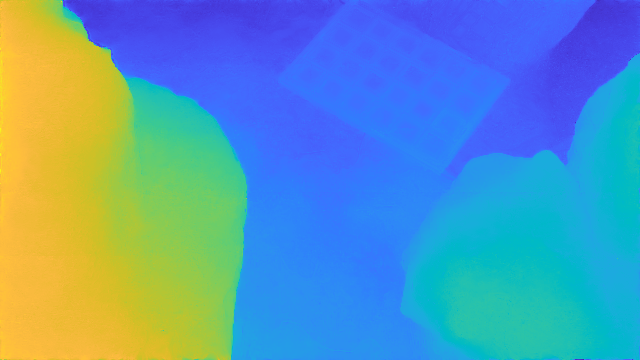}\hfill
    \includegraphics[width=0.16\linewidth]{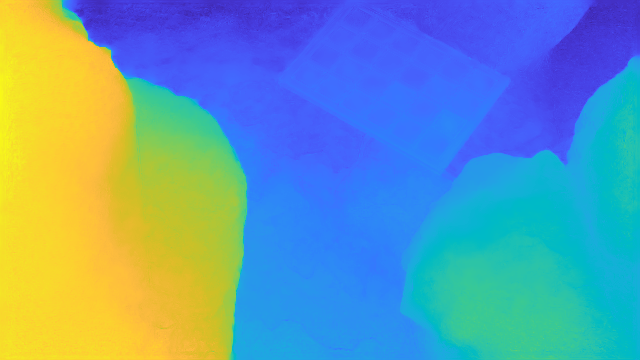}\hfill
    \includegraphics[width=0.16\linewidth]{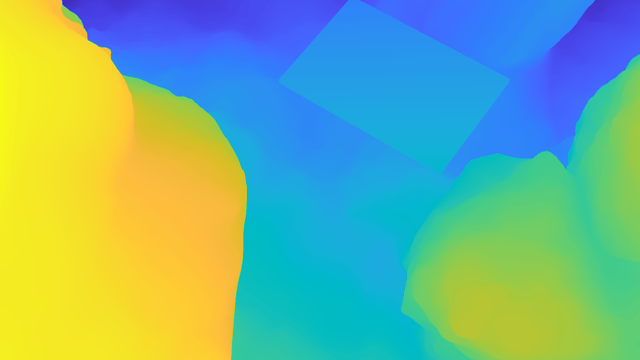} \\[4pt]
    \includegraphics[width=0.16\linewidth]{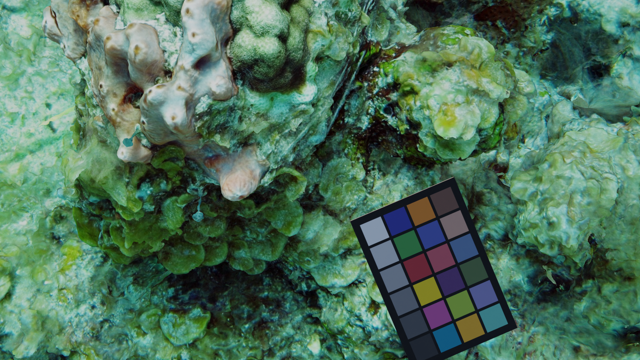}\hfill
    \includegraphics[width=0.16\linewidth]{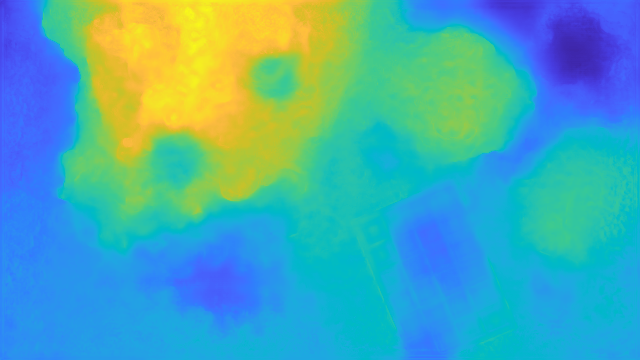}\hfill
    \includegraphics[width=0.16\linewidth]{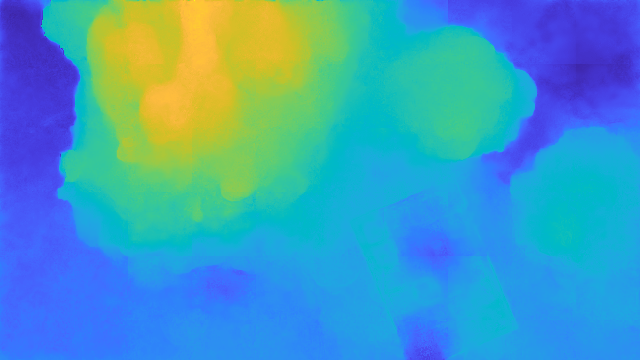}\hfill
    \includegraphics[width=0.16\linewidth]{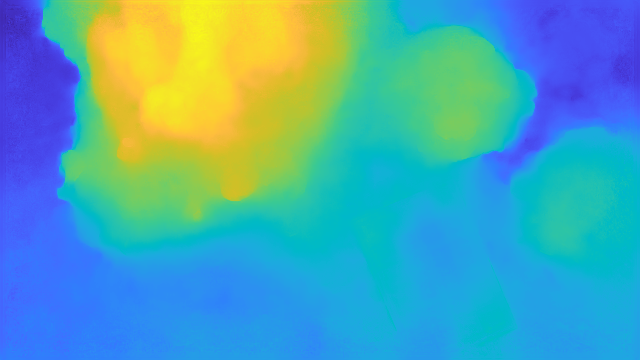}\hfill
    \includegraphics[width=0.16\linewidth]{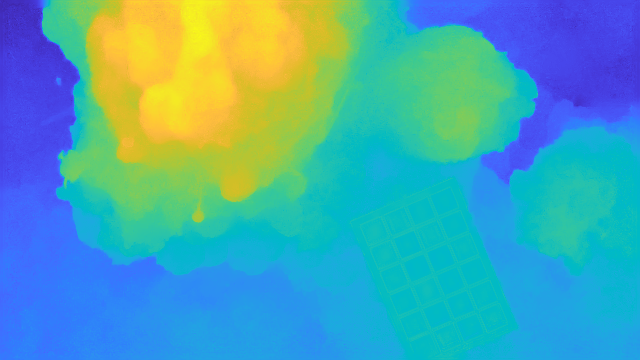}\hfill
    \includegraphics[width=0.16\linewidth]{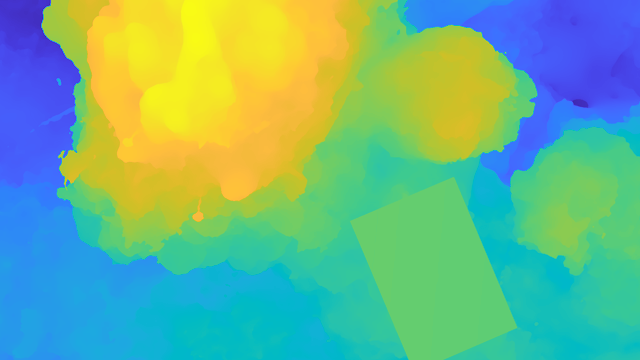}
    
    \small
    \makebox[0.16\linewidth]{(a) GT Image} \hfill
    \makebox[0.16\linewidth]{(b) DistgSSR~\cite{wang2022disentangling}} \hfill
    \makebox[0.16\linewidth]{(c) MSPNet~\cite{wang2023multi}} \hfill
    \makebox[0.16\linewidth]{(d) LFUB~\cite{lin2025enhancing}} \hfill
    \makebox[0.16\linewidth]{(e) Ours} \hfill
    \makebox[0.16\linewidth]{(f) GT Depth}
    \caption{Visual comparisons of the center depth maps derived by applying the LF depth estimation method in~\cite{jin2022occlusion} to the enhanced LF images by different LF-based methods. Note that the GT depth is rendered from the LFUB dataset.}
    \label{fig:result-depth}
    \vspace{-3mm}
\end{figure*}

\subsubsection{Methods under Comparison}
To ensure sufficiency, we compared our method with various image/video/LF-based enhancement methods. These methods include three 2-D RGB physical model-based approaches, Fusion~\cite{ancuti2012enhancing}, GDCP~\cite{peng2018generalization}, WWPF~\cite{zhang2023underwater}, and five 2-D RGB data-driven methods, LANet~\cite{liu2022adaptive}, PUIE~\cite{fu2022uncertainty}, Ushape~\cite{peng2023u}, Uranker~\cite{guo2023underwater}, UIEDP~\cite{du2025uiedp}, and one video-based enhancement method, UVE~\cite{du2024end}, and three 4-D LF-based enhancement methods, DistgSSR~\cite{wang2022disentangling}, MSPNet~\cite{wang2023multi}, LFUB\footnote{We use ``LFUB'' to denote both the method and the dataset mentioned in~\cite{lin2025enhancing}.}~\cite{lin2025enhancing}. For training and testing 2-D RGB-based methods, each sub-aperture image (SAI) of the LF images was treated as an individual input. For the video-based method, we selected several SAIs from adjacent viewpoints to form a single input sequence. All comparative evaluations utilized the author-released codes to train and present results on the LFUB dataset.

\subsubsection{Evaluation Metrics}
We employed widely used metrics to quantitatively evaluate the performance of our method and compare it with other methods. For paired LFUB dataset testing, we utilized peak signal-to-noise ratio (PSNR), structural similarity (SSIM)~\cite{Wang2004ImageQA}, learned perceptual image patch similarity (LPIPS)~\cite{Zhang2018TheUE}, and DeltaE (CIE2000 standard, $\vartriangle$$E$)~\cite{sharma2005ciede2000} metrics, which are full-reference metrics calculated based on the no-water images in the LFUB dataset for a fair comparison with existing methods. Note that DeltaE, as commonly applied in enhancement tasks, is used to evaluate color differences for enhancement performance. We also evaluated all algorithms on the newly released LFUID\footnote{We use ``LFUID'' to denote both the method and the dataset mentioned in~\cite{zhou2025lfuid}. We have extracted the experimental results provided in the original paper in Table~\ref{tab-exp-real}. However, since the source code has not been released, we are unable to provide visual comparison results.} dataset and compared with its LF-based method~\cite{zhou2025lfuid}, which features a large amount of real-world captured scenes. To evaluate on this real-world dataset, we mainly focus on the non-reference metrics, including underwater image quality measure (UIQM)~\cite{panetta2015human}, blind/referenceless image spatial quality evaluator (BRISQUE)~\cite{mittal2012no}, neural image assessment (NIMA)~\cite{talebi2018nima}, and colorfulness contrast fog density (CCF)~\cite{wang2018imaging}. Moreover, to evaluate the ability of preserving the geometric structure in LF images, we compared the depth maps derived by the LF depth estimation method~\cite{jin2022occlusion} from the enhanced LF images.

\begin{figure*}[t]
\centering
\subfloat[Input]{%
  \includegraphics[width=0.24\linewidth]{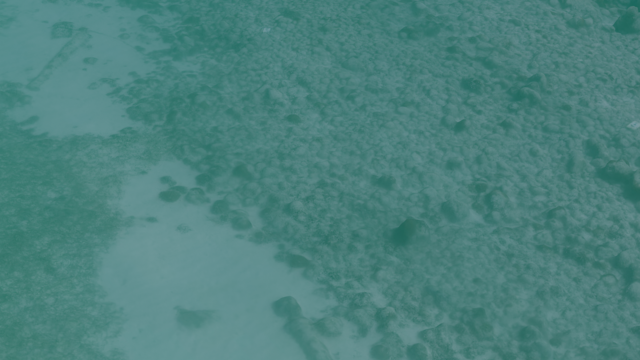}
}
\hfil
\subfloat[w/o Conv-adapter]{%
  \includegraphics[width=0.24\linewidth]{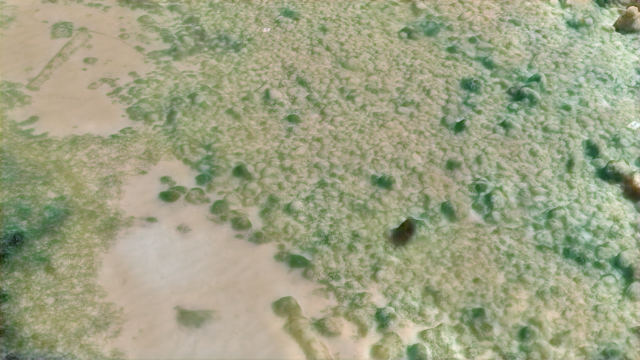}
}
\hfil
\subfloat[w/o EPI-adapter]{%
  \includegraphics[width=0.24\linewidth]{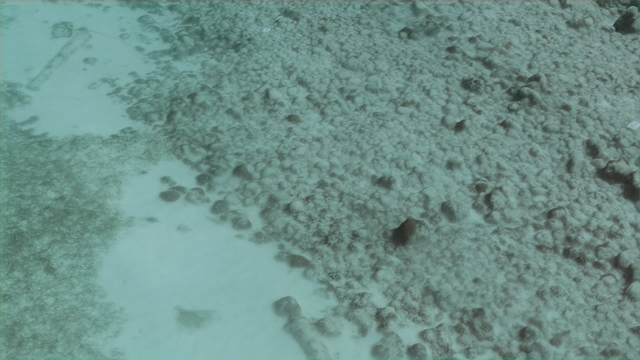}
}
\hfil
\subfloat[Ours]{%
  \includegraphics[width=0.24\linewidth]{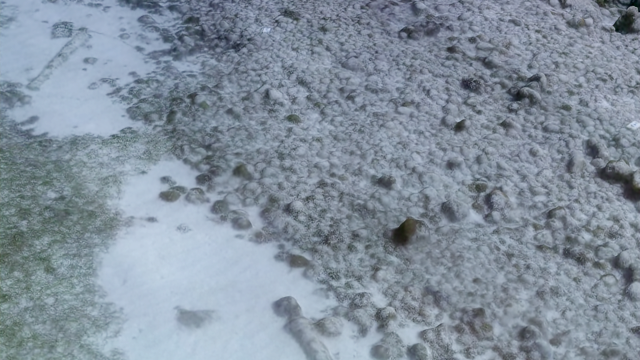}
}
\hfil

\caption{Visual comparisons of ablation studies for the impact of two Adapters.}
\label{fig:result-3}
\end{figure*}

\subsection{Comparison with State-of-the-Art Methods}
\subsubsection{Results on Paired LF Image Dataset} 

We first test the effectiveness of the proposed method on the paired LFUB dataset. The quantitative results in terms of PSNR, SSIM, LPIPS, and DeltaE of each competing method are summarized in Table~\ref{tab-results}, from which it can be found that our method consistently demonstrates the best performance. Notably, our method achieves higher performance than LFUB, the first 4-D LF-based underwater enhancement method, across all four metrics, making it better able to address LF-based problems. Obviously, our method achieves a 1.8-point improvement in DeltaE, demonstrating a particularly pronounced advantage in restoring color balance. 
Fig.~\ref{fig:result-1} and Fig.~\ref{fig:result-2} visually compare the enhanced SAIs and EPI slices of all approaches across two scenes with bluish and greenish color deviations. As can be observed, the visual quality of each method is consistent with the numerical results reported in Table~\ref{tab-results}. Specifically, as shown in Figs.~\ref{fig:result-1},  Ushape~\cite{peng2023u}, DistgSSR~\cite{wang2022disentangling}, and MSPNet~\cite{wang2023multi} are observed to cause a noticeable color shift. In contrast, our method excels at restoring colors in the red-words region of Fig.~\ref{fig:result-1} without extra color artifacts in the visual results of UVE~\cite{du2024end}, UIEDP~\cite{du2024end}, and LFUB~\cite{lin2025enhancing}. Compared to these methods, our superior performance is attributable to the regularity of the geometry regularization we designed. As shown in Fig.~\ref{fig:result-2}, our method successfully eliminates the greenish color deviation while maintaining consistent color balance, whereas other methods tend to suffer from local strong color casts. The competing methods (especially 2-D RGB-based and video-based approaches) tend to produce blurred, discontinuous, or bent EPI lines, indicating corrupted disparity and cross-view inconsistency. In contrast, GeoDiff-LF maintains sharp, straight, and coherent epipolar lines even in severely degraded underwater scenes, which directly demonstrates that the proposed global geometry regularization and EPIT-Adapter jointly preserve correct LF geometric structure. It is worth noting that the 2-D diffusion-based method UIEDP~\cite{du2025uiedp} outperforms other single-image approaches, indicating that incorporating diffusion priors is indeed beneficial for underwater image enhancement. Nevertheless, their performance remains inferior to ours. We attribute this gap to the diffusion framework's failure to account for the intrinsic structural properties of light fields, thereby failing to preserve geometric consistency during restoration.

\subsubsection{Results on Real-Captured LF Image Dataset} 
To demonstrate the generalization ability, we applied several methods (i.e., WWPF~\cite{zhang2023underwater}, Ushape~\cite{peng2023u}, Uranker~\cite{guo2023underwater}, LFUB~\cite{lin2025enhancing}, LFUID~\cite{zhou2025lfuid}, and ours) trained on the LFUB dataset to the real captured underwater LF dataset LFUID. The enhancement performance, measured by UIQM, BRISQUE, NIMA, and CCF, is summarized in Table~\ref{tab-exp-real}. As shown, our method yields better UIQM, BRISQUE, and NIMA scores than all competitors, demonstrating stronger generalization. 
While it trails model-based methods such as WWPF~\cite{zhang2023underwater} and LFUID~\cite{zhou2025lfuid} on the CCF metric, which primarily assesses fog removal, the performance gap stems mainly from the diffusion model treating larger particles in turbid water as fine details, inadvertently generating a residual fog layer that lowers CCF scores. We also test two underwater LF image enhancement methods, LFUB~\cite{lin2025enhancing} and ours, on four common real-world underwater degradation types in the LFUID dataset in Fig.~\ref{fig:result-real}, which confirm the numerical results in Table~\ref{tab-exp-real}. Specifically, as shown in Fig.~\ref{fig:result-real}, our model produces crisper details and more well-balanced colors. This observation is further illustrated in Figs.~\ref{fig:result-real}(c)(f), where enhanced particles contribute to the residual fog layer, directly explaining the reduced CCF scores. Overall, these results highlight the strong potential of our approach for practical underwater imaging applications.

\begin{table}[t]
\centering
\caption{Quantitative results of ablation studies assessing the impact of two Adapters within our proposed network, such as the Multi-pattern Convolutional Adapter and the EPI-guided Attention Adapter. ``\checkmark'' (resp. blank) represents the corresponding structure is used (resp. unused)}
\label{tab-abfeature}
\setlength{\tabcolsep}{4mm}{
\small
\begin{tabular}{c|ccc}
\toprule[1.2pt]
 & (a) & (b) & (c)  \\ 
\midrule
Conv-Adapter       & \checkmark &            & \checkmark \\ 
EPI-Adapter        &            & \checkmark &  \checkmark \\ 
\midrule
PSNR $\uparrow$    & 21.43  & 21.05  & \textbf{22.51} \\ 
SSIM $\uparrow$    & 0.8611 & 0.8490 & \textbf{0.8699} \\ 
% LPIPS $\downarrow$ & 0.2613 & 0.2682 & \textbf{0.2581} \\ 
\bottomrule[1.2pt]
\end{tabular}}
\vspace{-3mm}
\end{table}

\subsubsection{Depth Estimation Results on the Enhanced LF Images} 
Depth estimation in underwater scenes is prone to bias, and the accuracy of depth maps derived from LF images depends heavily on the quality of the enhanced images. To evaluate this aspect, we compare the depth estimation performance across LF-based methods (i.e., DistgSSR~\cite{wang2022disentangling}, MSPNet~\cite{wang2023multi}, LFUB~\cite{lin2025enhancing}, and ours) using an unsupervised LF depth estimation approach~\cite{jin2022occlusion}. As illustrated in Fig.~\ref{fig:result-depth}, our method produces depth maps that are closest to the ground truth, with sharper edges of occlusion boundaries and better preservation of smoothness in uniform-depth regions. These results highlight the superior ability of our approach to maintain geometric structure in enhanced LF images compared to competing methods.

\begin{table}[t]
\caption{Quantitative results of ablation study about different regularization, including L1, L2, SSIM and our global geometry regularization.}
\label{tab-exp-regularization}
\centering
\small
\setlength{\tabcolsep}{4mm}{
\begin{tabular}{c|cccc}
\toprule[1.2pt]
Loss & L1  & L2 & SSIM & Ours \\ 
\midrule
PSNR $\uparrow$       & 21.15 & 21.52 & 21.35 & \textbf{22.51}  \\
SSIM $\uparrow$       & 0.8474 & 0.8541 & 0.8482 & \textbf{0.8699} \\
% LPIPS $\downarrow$    & 0.2823 & 0.2784 & 0.2805 & \textbf{0.2581} \\
\bottomrule[1.2pt]
\end{tabular}}
\vspace{-3mm}
\end{table}

\begin{table}[t]
\caption{Quantitative results of the ablation study about the placement of regularization between variables.}
\label{tab-place}
\centering
\small
\setlength{\tabcolsep}{3mm}{
\begin{tabular}{c|ccc}
\toprule[1.2pt]
Placement & btw. $\mathbf{\hat{X}}$ and $\mathbf{X}_{0}$ & btw. $\mathbf{X}_{\tau}$ and $\mathbf{X}_{0}$ & Ours \\ 
\midrule
PSNR $\uparrow$       & 21.53 & 21.76 & \textbf{22.51}  \\
SSIM $\uparrow$       & 0.8545 & 0.8589 & \textbf{0.8699} \\
% LPIPS $\downarrow$    & 0.2766 & 0.2659 & \textbf{0.2581} \\
\bottomrule[1.2pt]
\end{tabular}}
\vspace{-3mm}
\end{table}

\begin{table*}[t]
\scriptsize
  \caption{Comparisons of inference time (in seconds) by different methods on the LFUB dataset.}
  \label{tab-inf}
  \centering
  \small
  \setlength{\tabcolsep}{2mm}{
  \begin{tabular}{c|cccccc}
    \toprule[1.2pt]
    ~~~Methods~~~ 
    &~Fusion~\cite{ancuti2012enhancing}~
    &~GDCP~\cite{peng2018generalization}~
    &~WWPF~\cite{zhang2023underwater}~
    &~LANet~\cite{liu2022adaptive}~ 
    &~PUIE~\cite{fu2022uncertainty}~ 
    &~Ushape~\cite{peng2023u}~ \\
    
    \midrule
    Time
    &0.3512 &0.6469 &0.7557 &0.1122 &0.2823 &0.1836        \\ 
    
    \midrule[1.2pt]
     ~~~Methods~~~  
    &~Uranker~\cite{guo2023underwater}~
    &~UVE~\cite{du2024end}~
    &~DistgSSR~\cite{wang2022disentangling}~
    &~MSPNet~\cite{wang2023multi}~
    &~LFUB~\cite{lin2025enhancing}~
    &~Ours~ \\
    \midrule
    Time 
    &0.1727 &0.2449 &1.0269 &0.9196 &0.6987  &2.7952     \\ 
    \bottomrule[1.2pt]
  \end{tabular}}
  \vspace{-3mm}
\end{table*}

\begin{figure*}
\centering
\subfloat[Input]{%
  \includegraphics[width=0.24\linewidth]{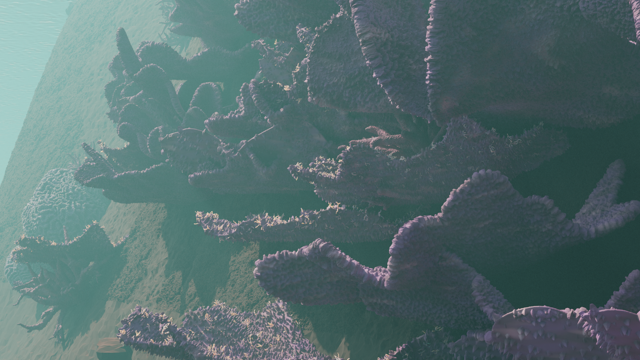}
}
\hfil
\subfloat[LFUB~\cite{lin2025enhancing}]{%
  \includegraphics[width=0.24\linewidth]{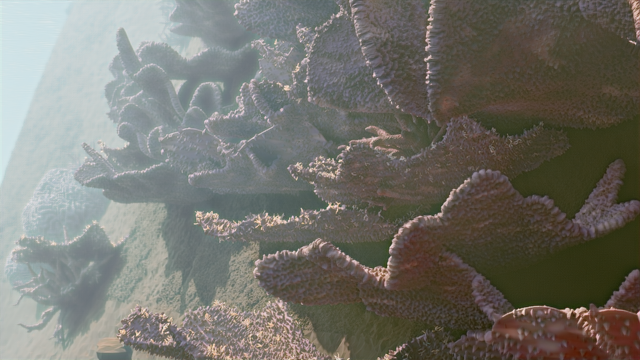}
}
\hfil
\subfloat[Ours]{%
  \includegraphics[width=0.24\linewidth]{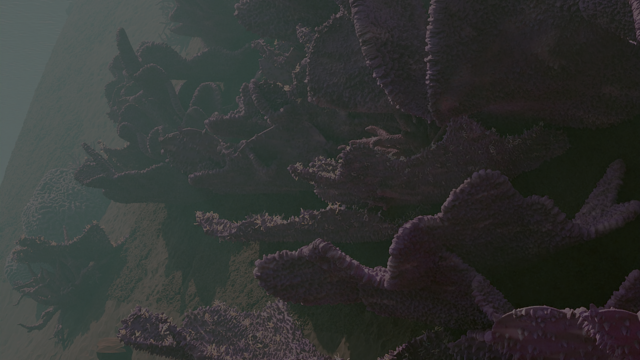}
}
\hfil
\subfloat[GT]{%
  \includegraphics[width=0.24\linewidth]{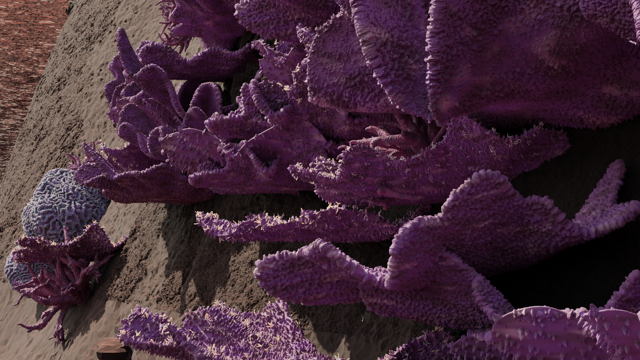}
}
\hfil

\caption{Visual results of one challenge case.}
\label{fig:result-challenge}
\vspace{-3mm}
\end{figure*}

\subsection{Ablation Study}
\label{sec:ablation}
To better understand the whole mechanism of the proposed GeoDiff-LF, we further conduct a series of ablation studies. Specifically, we investigate how the components of our method affect the performance of enhancing 5~$\times$~5 underwater LF images, including our lightweight adapters, the proposed global geometry regularization, and the computational efficiency.

\subsubsection{Evaluation of Lightweight Adapters}
As we proposed in our adapted diffusion architecture, the lightweight adapters contain many benefits during addressing 4-D LFs, as shown in Table~\ref{tab-abfeature} and Fig.~\ref{fig:result-3}. Obviously, by incorporating adapters into the pre-trained diffusion prior and fine-tuning only those adapters, we avoid the substantial computational burden of training a full diffusion model from scratch. This approach enables the proposed model to adaptively extract structural information from 4-D LF images while preserving the original diffusion prior knowledge. In Table~\ref{tab-abfeature}, the entire network achieves better enhancement performance than the ablated models on the test set, suggesting the effectiveness of both components. Furthermore, as depicted in Fig.~\ref{fig:result-3}, two adapters together can help balance the greenish color cast of the input image. This successful balance corroborates the superior efficacy of our proposed method.

\subsubsection{Evaluation of Global Geometry Regularization}
We examined the impact of various regularization strategies applied between the intermediate reconstruction $ \mathbf{X}_{t-1} $ and the clean target $ \mathbf{X}_0 $ on the diffusion framework's performance. As reported in Table~\ref{tab-exp-regularization}, our global geometry regularization yields substantially greater improvements compared to conventional pixel-wise losses (e.g., L1, L2, and SSIM). This advantage arises from that standard losses cannot effectively capture inter-view dependencies or recurring geometric patterns in 4-D LF images, whereas our approach explicitly models these global relationships. These findings clearly underscore the effectiveness of global geometry regularization in maintaining holistic geometric consistency.

\subsubsection{Evaluation of Intermediate Reconstruction}
We delved into the impact of optimal placement of regularization between variables, including between $\mathbf{X}_{\tau}$ and $\mathbf{X}_{0}$, and between $\mathbf{\hat{X}}$ and $\mathbf{X}_{0}$, on the enhancement of diffusion models. The quantitative results in Table~\ref{tab-place} demonstrate that our stepwise fusion strategy achieves the best performance. We attribute this superiority to the progressive nature of the proposed approach, which consistently refines the reconstruction toward the ground-truth images at each step. This iterative refinement process may also help mitigate the impact of cumulative errors that inevitably arise in intermediate reconstructions due to error propagation across timesteps.

\subsubsection{Evaluation of Timestep Schedule}
We conduct ablations on both the training starting timestep and the inference schedule. As shown in Table~\ref{tab-abtrainst}, training with $\tau \in \{500,400,300,200\}$ achieves the best results. Starting from 600 degrades performance because excessive noise destroys geometric cues, while starting from 400 alone is insufficient because the model misses the high-noise regime essential for robust cross-step denoising. Table~\ref{tab-absamplest} compares multiple schedules. Our 5-step schedule $\{500,400,300,200,100\}$ performs best. The sparse 3-step schedule $\{500,300,100\}$ remains competitive, confirming the noise predictor's effectiveness. However, single-step sampling (e.g., $\{500\}$ or $\{100\}$) causes clear degradation, indicating that a moderate number of transitions is necessary for stable refinement. These results validate that our selected schedules strike an effective balance between fidelity and efficiency.

\begin{table}[H]
\centering
\caption{Quantitative results of ablation studies on the starting timestep.}
\label{tab-abtrainst}
\setlength{\tabcolsep}{4mm}{
\small
\begin{tabular}{c|ccc}
\toprule[1.2pt]
\#Steps & PSNR & SSIM  \\ 
\midrule
$\{600,500,400,300,200\}$   & 21.73           & 0.8612 \\ 
$\{500,400,300,200\}$       & \textbf{22.51} & \textbf{0.8699}           \\ 
$\{400,300,200\}$       & 21.58           & 0.8573 \\ 
\bottomrule[1.2pt]
\end{tabular}}
\end{table}

\begin{table}[H]
\centering
\caption{Quantitative results of ablation studies on the timestep schedule.}
\label{tab-absamplest}
\setlength{\tabcolsep}{4mm}{
\small
\begin{tabular}{c|ccc}
\toprule[1.2pt]
\#Steps & PSNR & SSIM  \\ 
\midrule
$\{500,400,300,200,100\}$       & \textbf{22.51} & \textbf{0.8699}           \\ 
$\{500,300,100\}$       & 22.29           & 0.8614 \\ 
$\{400,250,100\}$        & 22.10           & 0.8579 \\ 
$\{300,200,100\}$        & 21.98           & 0.8583 \\ 
$\{500\}$        & 22.11           & 0.8605 \\ 
$\{300\}$        & 21.54           & 0.8574 \\ 
$\{100\}$        & 21.46           & 0.8558 \\ 
\bottomrule[1.2pt]  
\end{tabular}}
\end{table}

\subsubsection{Evaluation of Computational Efficiency} 
In this study, we conducted a comparative analysis of the inference time of various methods on the LFUB dataset, as presented in Table~\ref{tab-inf}. The reported inference time for 2-D RGB and video-based methods represents the estimated processing time per individual image or per frame. As observed, our method incurs a slightly longer inference time compared to other LF-based approaches, primarily due to the multi-step sampling process. Nevertheless, the overall runtime remains within an acceptable range relative to existing methods. This observation motivates our future work to further reduce the number of sampling steps while maintaining enhancement performance, thereby improving efficiency.

\subsection{Challenging Case and Limitations}\label{sec:limitation}
While our proposed method demonstrates notable performance gains, as evidenced in prior sections, certain challenging scenarios remain prevalent in underwater LF enhancement. Fig.~\ref{fig:result-challenge} presents a representative case where frequent and abrupt color variations in specific regions hinder both our approach and competing methods from achieving visually pleasing color restoration. Such results can be partially attributed to the fact that currently, we do not explicitly model underwater physical processes (such as wavelength-dependent attenuation or backscatter) \cite{wang2017single}, which leads to our method struggling in extreme scenarios where the degradation violates the training distribution, such as scenes with highly turbid water or severe color fading. In addition, the current framework adopts a fixed starting timestep determined from the overall training set statistics. While the noise map predictor $f_w$ adapts its output to the input content, the timestep itself does not vary with degradation severity. Consequently, under highly turbid conditions where the image structure is severely corrupted, the fixed schedule may cause the predicted noise to deviate from the ideal initialization, yielding suboptimal results.

To address these limitations, future work will focus on integrating explicit underwater physical prior to further enhance robustness in extreme turbidity and severe color-fading scenarios, and exploring content-adaptive strategies that dynamically adjust the sampling trajectory based on estimated degradation severity would improve generalization across diverse underwater environments.

\section{Conclusion}
\label{sec:c}

In this paper, we propose a novel diffusion-based framework, GeoDiff-LF, for underwater LF image enhancement, leveraging the geometric information of 4-D LF to address depth-dependent degradations, such as absorption and scattering. Our method introduces convolutional and attention adapters for geometric modeling, a global geometry regularization for structure preservation, and noise prediction for efficient sampling. This achieves superior restoration quality compared to all other methods. 

In the future, we will devote ourselves to integrating explicit underwater physical prior to the proposed framework, and exploring content-adaptive strategies that dynamically adjust the sampling trajectory based on estimated degradation severity, as discussed in \ref{sec:limitation}, thereby better meeting the practical requirements of real-world underwater imaging applications.

\bibliographystyle{IEEEtran}
\bibliography{ref}

\begin{IEEEbiography}[{\includegraphics[width=1in, height=1.25in, clip, keepaspectratio]{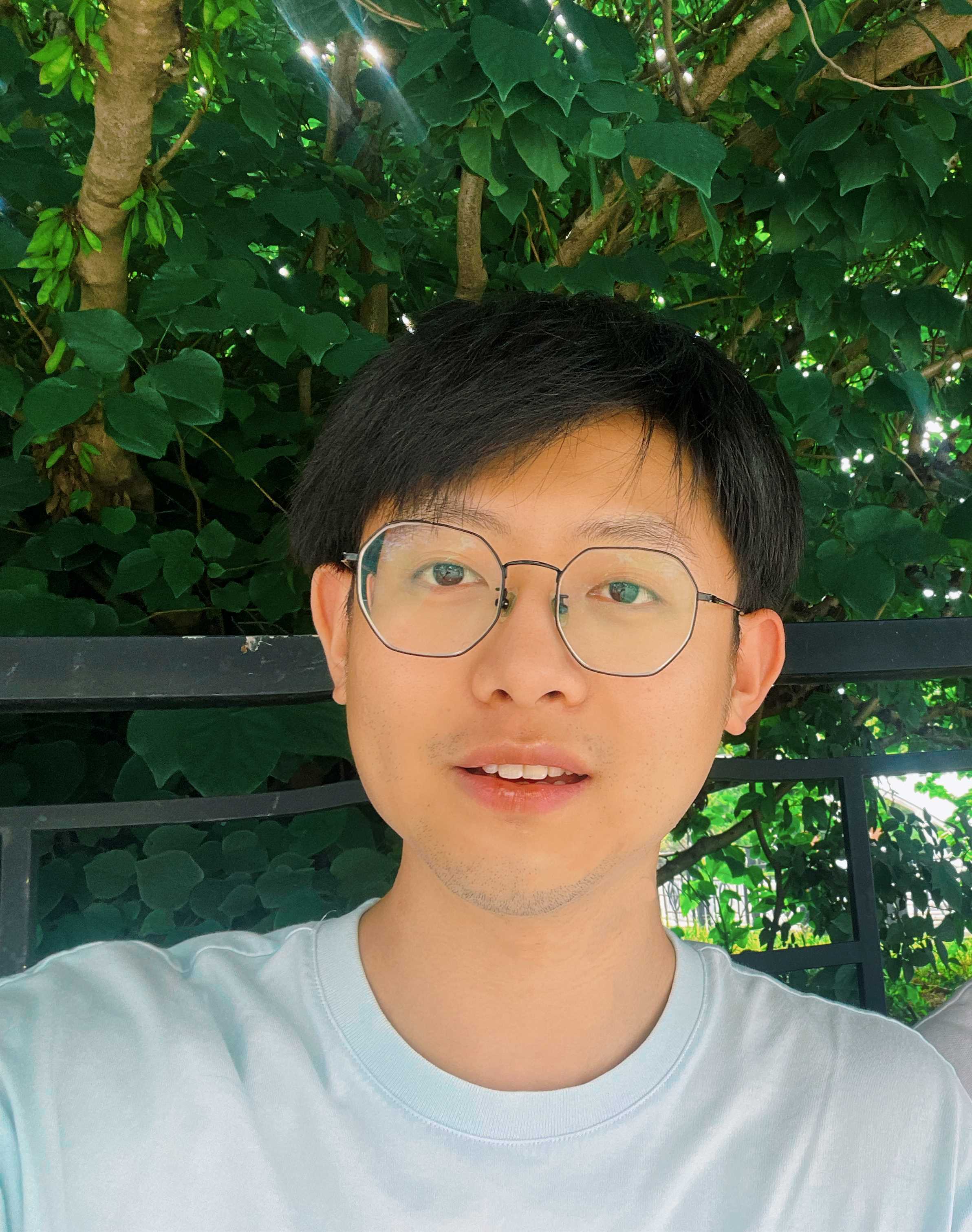}}]{Yuji Lin} received the B.Sc. and Ph.D. degrees from the School of Mathematics and Statistics, Xi'an Jiaotong University, Xi'an, China, in 2020 and 2026, respectively.

He was a Visiting Scholar with City University of Hong Kong, Hong Kong SAR, from 2023 to 2024.
His research interests include light field imaging, image restoration, and model-based deep learning.
\end{IEEEbiography}

\begin{IEEEbiography}[{\includegraphics[width=1in,height=1.25in, clip,keepaspectratio]{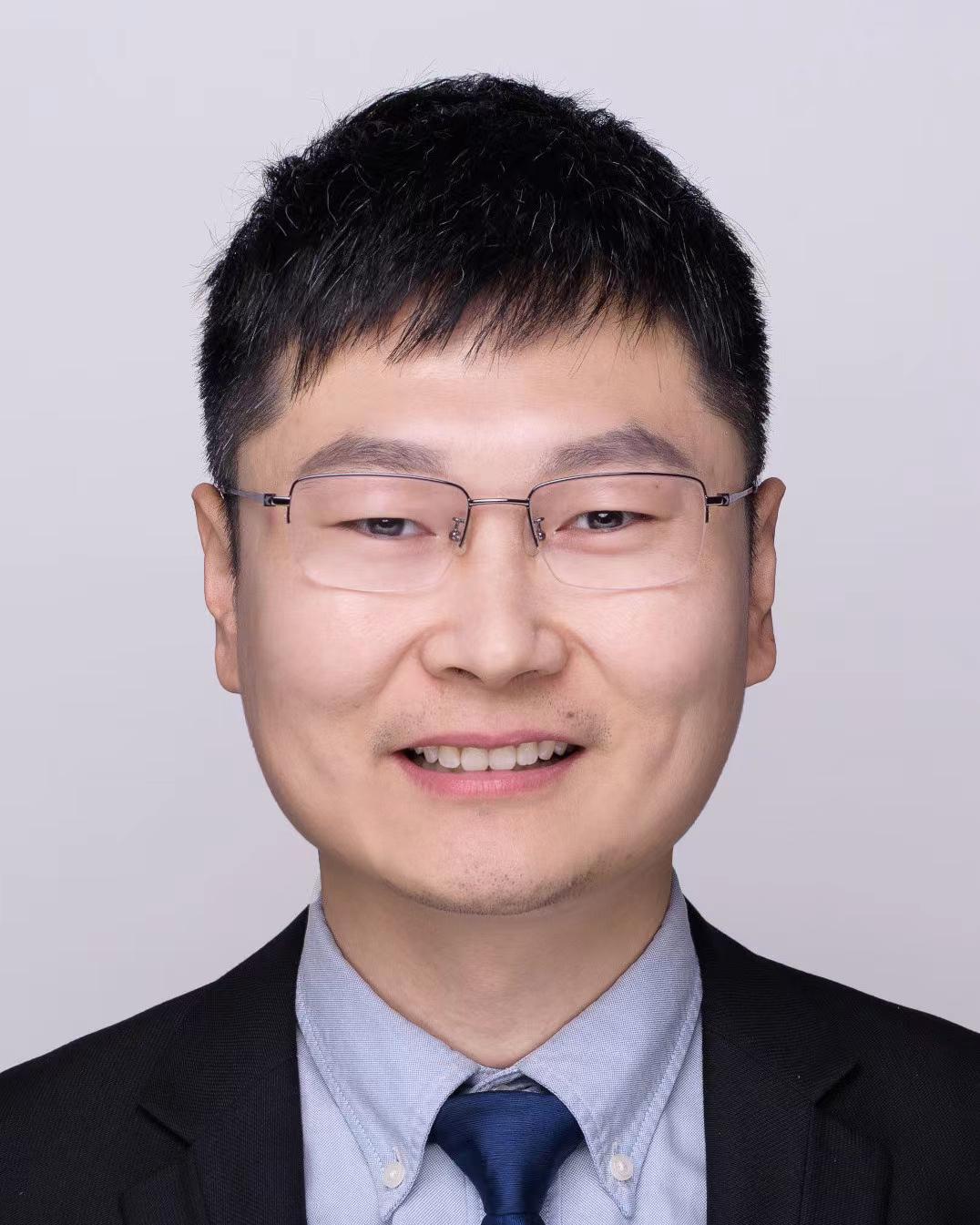}}]{Qian Zhao} (Member, IEEE) received the B.Sc. and Ph.D. degrees from Xi’an Jiaotong University, Xi’an,
China, in 2009 and 2015, respectively. 

He was a Visiting Scholar with Carnegie Mellon University, Pittsburgh, PA, USA, from 2013 to 2014. He is currently a Professor with the School of Mathematics and Statistics, Xi’an Jiaotong University. His research interests include low-rank matrix/tensor analysis, Bayesian modeling, and meta-learning.
\end{IEEEbiography}

\begin{IEEEbiography}[{\includegraphics[width=1in, height=1.25in, clip, keepaspectratio]{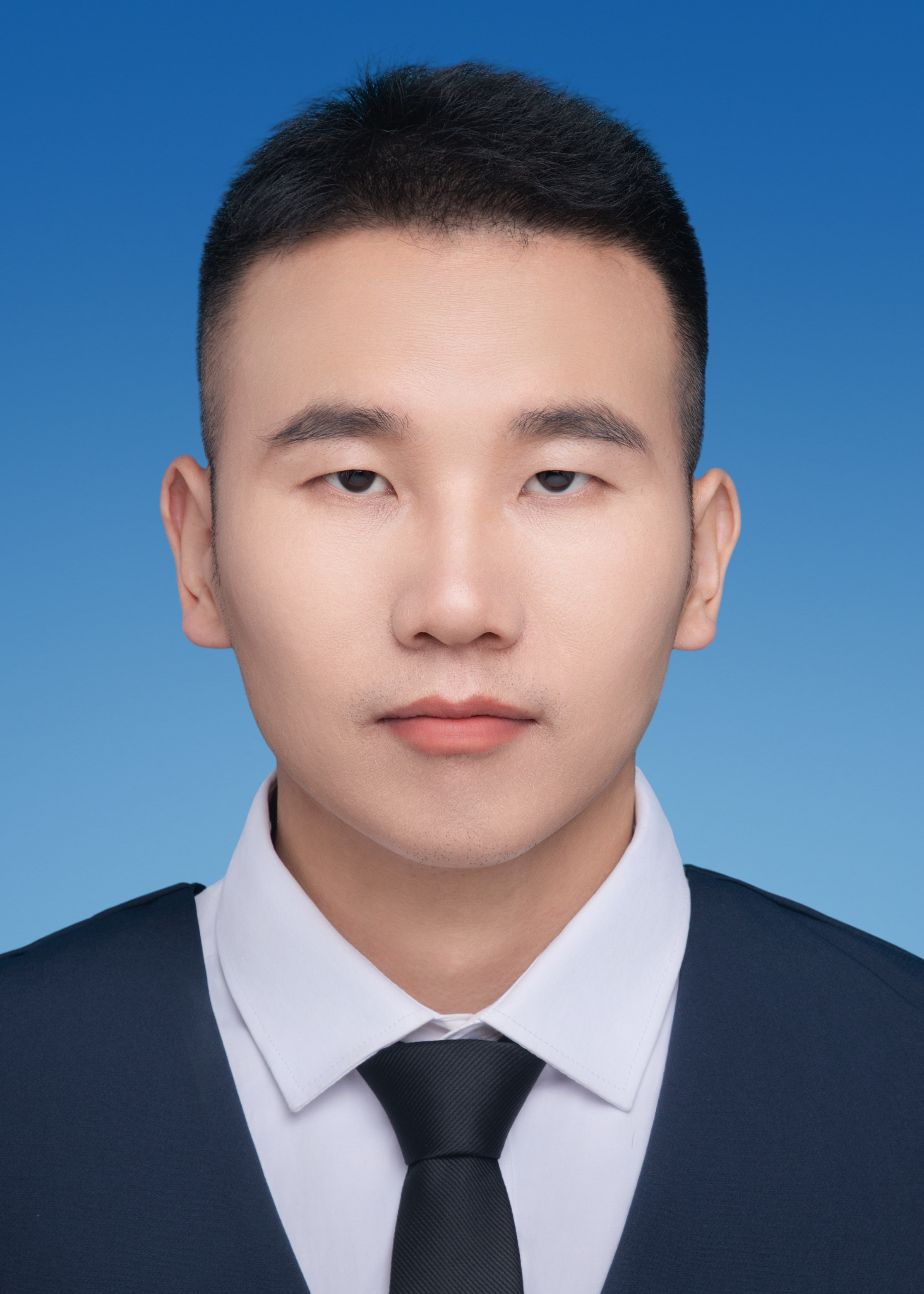}}]{Zongsheng Yue} (Member, IEEE) received the Ph.D. degree from Xi’an Jiaotong University, Xi’an, China, in 2021. 

He was a Postdoctoral Research Fellow with the School of Computer Science and Engineering, Nanyang Technological University, Singapore, from 2022 to 2025, and the Department of Computer Science, Hong Kong University, Hong Kong, from 2021 to 2022. He is a Professor with the School of Mathematics and Statistics, Xi’an Jiaotong University. His research interests include noise modeling, image restoration, and diffusion models.
\end{IEEEbiography}

\begin{IEEEbiography}[{\includegraphics[width=1in, height=1.25in, clip, keepaspectratio]{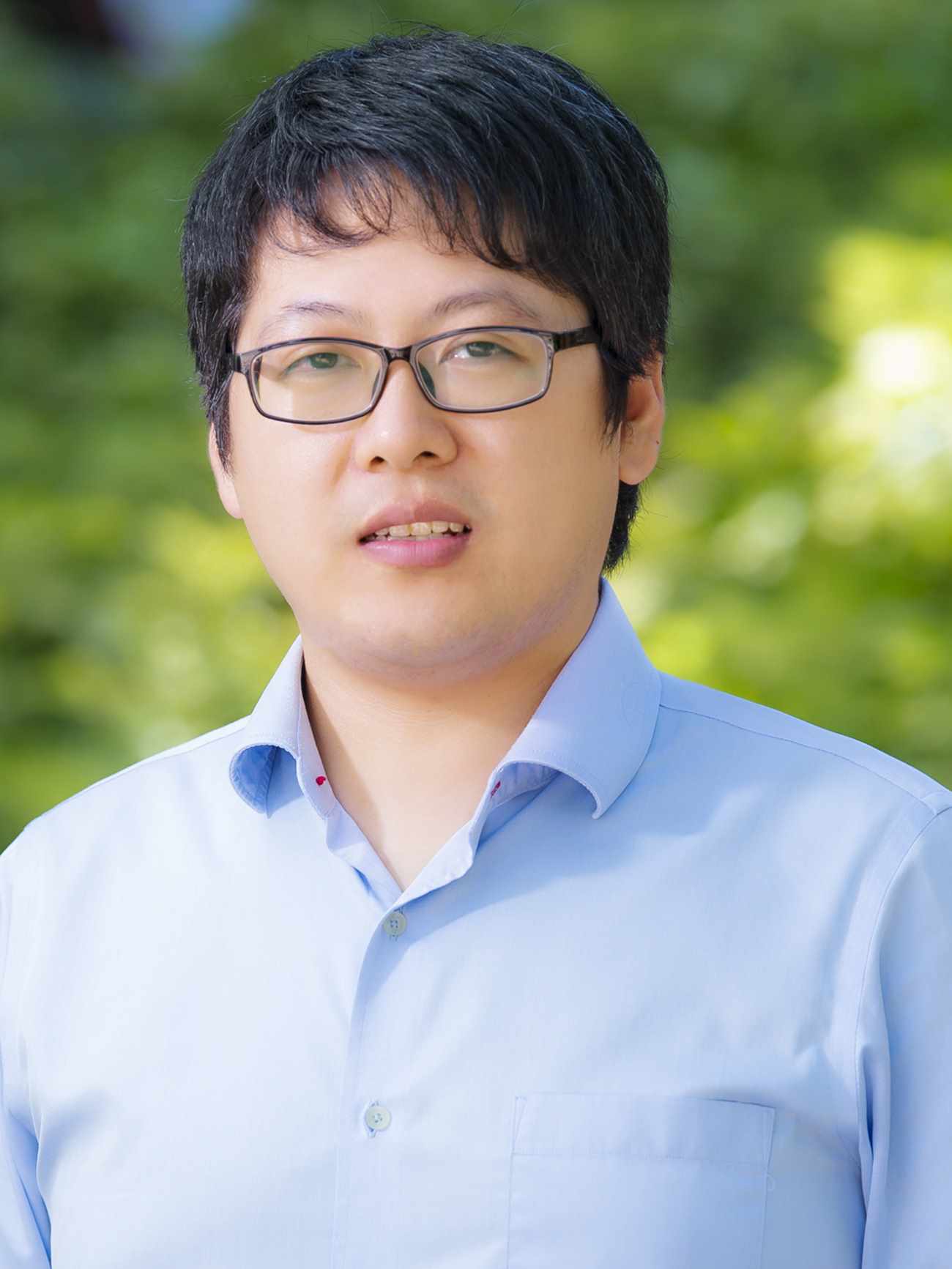}}]{Junhui Hou} (Senior Member, IEEE) is a Professor with the Department of Computer Science, City University of Hong Kong. His research interests include multidimensional visual computing, such as light field, hyperspectral, geometry, and event data.

He received the Early Career Award from the Hong Kong Research Grants Council in 2018, IEEE Multimedia Rising Star Award in 2023, the Excellent Young Scientists Fund from NSFC in 2024, and the IEEE TIP Best Paper Award in 2025. He is serving as a Senior Area Editor for IEEE TIP and an Associate Editor for IEEE TVCG and TMM. He served as an Associate Editor for IEEE TIP and TCSVT. 
\end{IEEEbiography}

\begin{IEEEbiography}[{\includegraphics[width=1in,height=1.25in, clip,keepaspectratio]{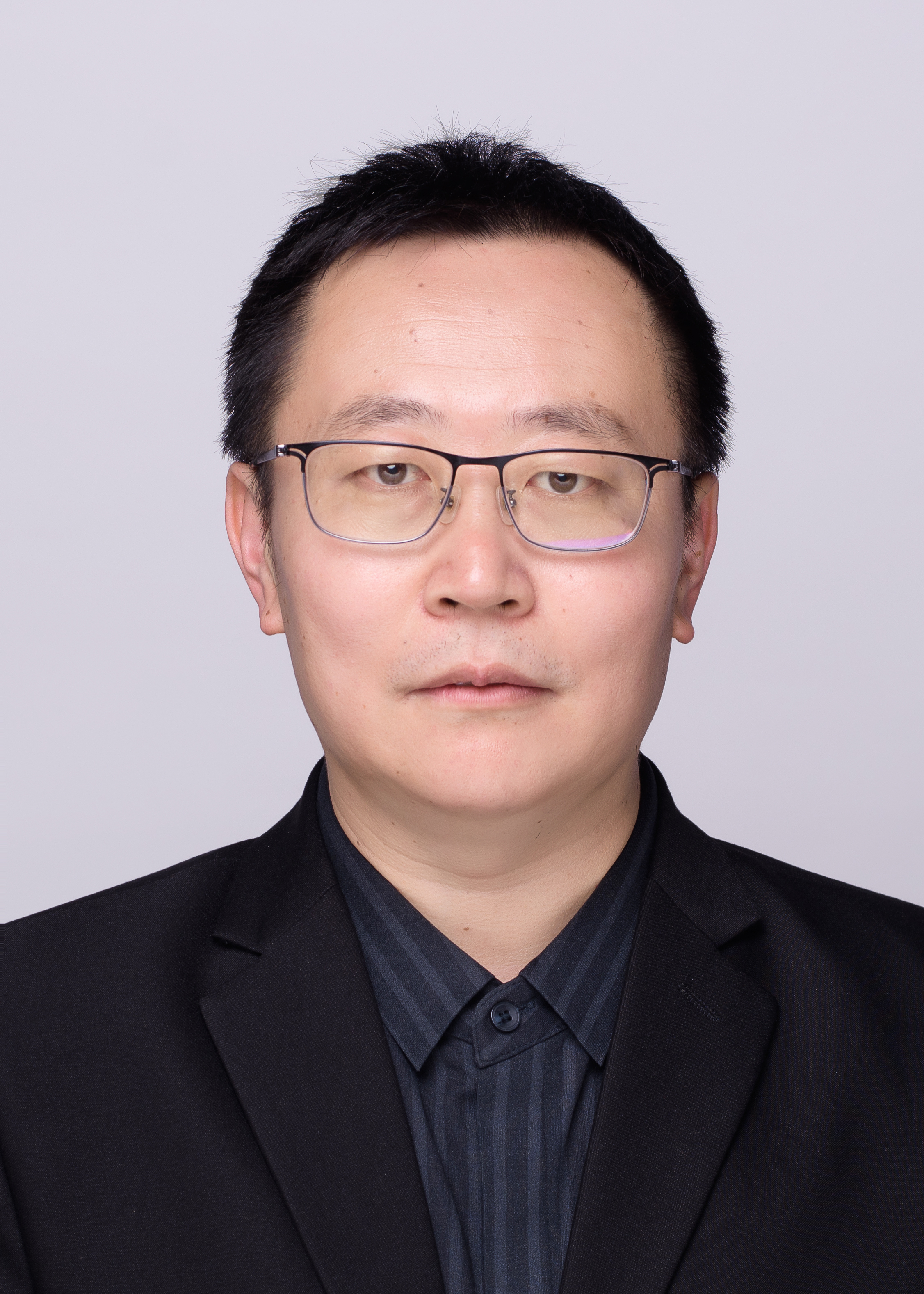}}]{Deyu Meng} (Member, IEEE) received the B.Sc., M.Sc., and Ph.D. degrees from Xi'an Jiaotong University, Xi'an, China, in 2001, 2004, and 2008, respectively. 

He was a Visiting Scholar with Carnegie Mellon University, Pittsburgh, PA, USA, from 2012 to 2014. He is currently a Professor with the School of Mathematics and Statistics, Xi’an Jiaotong University, and Adjunct Professor with the Faculty of Information Technology, Macau University of Science and Technology, Macau, China. His research interests include model-based deep learning, variational networks, and meta-learning.
\end{IEEEbiography}

\end{document}